**Field Validation of a Multi-Resolution ConvLSTM Framework for Retaining Wall Deformation Prediction**

Jihoon Kim, Saeyon Kim, and Heejung Youn*

Department of Civil and Environmental Engineering, Hongik University, Seoul 04066, Republic of Korea

* Corresponding author. E-mail address: geotech@hongik.ac.kr

**Abstract**

This study presents a comprehensive field validation of a multi-resolution Convolutional Long Short-Term Memory (ConvLSTM) framework for predicting retaining wall deformation during staged excavation. The framework is trained on Gaussian noise-augmented numerical simulations and integrates ConvLSTM models operating at different temporal resolutions through a stacking ensemble strategy. The proposed framework is validated using field monitoring data from 34 inclinometers across 11 excavation sites in South Korea. Site-wise prediction performance is systematically evaluated using multiple evaluation metrics, with analyses of the influence of temporal deformation irregularity and spatiotemporal prediction characteristics on model performance. The results demonstrate that the framework predicts retaining wall deformation associated with up to 5.0 m of additional excavation with an average mean absolute error of 1.4 mm and a coefficient of determination of 0.93 across the excavation sites. These results indicate that the framework, although trained exclusively on numerically simulated and augmented database, can be effectively applied to diverse field excavation conditions and achieve a reliable level of prediction accuracy in practical retaining wall deformation prediction.

## 1. Introduction

Deformation prediction has been a central topic in civil and geotechnical engineering, as excessive or unexpected structural displacement can directly compromise the serviceability and safety of infrastructure systems (Lim et al., 2018, Peck, 1969). Across a wide range of civil structures—including bridges, slopes, tunnels, and braced retaining wall systems—structural deformation evolves progressively under the combined influence of construction sequences, material nonlinearity, and soil–structure interaction (Chen et al., 2024, Hsiung, 2020, Huang, 2013, Zhang et al., 2023). Consequently, reliable prediction of future deformation is a fundamental requirement for construction management, risk mitigation, and early warning in engineering practice. Traditionally, deformation behavior has been estimated using empirical correlations and statistical approaches derived from accumulated case histories. Such methods have been widely applied to excavation-induced ground movements, slope deformation, tunnel settlement, and retaining wall behavior, providing useful reference values for maximum displacement and typical deformation envelopes (Clough, 1990, Haneberg, 2004, Long, 2001, Moorman, 2004). However, empirical approaches are inherently limited in capturing the spatial distribution and temporal evolution of deformation processes, particularly under staged construction and changing boundary conditions. As a result, these methods are mainly applicable for qualitative assessment or preliminary design and are insufficient for accurate, location-specific, and long-term deformation prediction.

To overcome these limitations, finite element method (FEM)–based numerical analysis and inverse analysis techniques have been extensively adopted in civil and geotechnical engineering. Numerical analysis enables physics-based simulation of construction sequences, stress redistribution, and soil–structure interaction under staged excavation, while inverse analysis calibrates model parameters to reproduce observed field responses by minimizing discrepancies between measured and simulated behavior (Houhou et al., 2019, Li et al., 2014). Despite their advantages, inverse analysis typically requires repeated nonlinear finite element simulations coupled with optimization algorithms, resulting in substantial computational cost and time. Although various optimization strategies and surrogate modeling techniques have been proposed to improve efficiency, inverse analysis remains difficult to implement in a timely manner during construction, particularly for real-time or near-real-time deformation prediction and decision-making (Gao, 2021, Zhao, 2024). Moreover, parameters identified through inverse analysis often depend strongly on modeling assumptions, boundary conditions, and monitoring locations, and may represent equivalent parameters that reproduce specific observations rather than intrinsic ground properties applicable to future construction stages.

With the increasing availability of long-term monitoring data, machine learning and deep learning techniques have gained attention as alternative approaches for deformation prediction. Previous studies have applied data-driven prediction models to a range of civil engineering problems (Du et al., 2025, Kim et al., 2025, Lai et al., 2025, Onyelowe et al., 2025) and have reported improved predictive performance compared with conventional statistical or empirical methods. These results suggest that data-driven and deep learning–based approaches are capable of capturing complex deformation behavior in civil engineering systems when sufficient monitoring data are available. Nevertheless, most existing studies have evaluated model performance under limited monitoring periods or site-specific conditions, leaving the robustness and general applicability of deep learning–based deformation prediction models across diverse and evolving construction environments insufficiently validated.

## 2. Related Works

Prediction of retaining wall deformation induced by staged excavation has received growing attention due to its critical role in construction safety and excavation management. In densely populated urban environments, the increasing demand for underground space has led to deeper and larger excavation projects, thereby intensifying the need for reliable and robust prediction of excavation-induced wall deformation. In response, numerous studies have explored data-driven and deep learning–based methods using either field monitoring data or numerical analysis results. Early studies on excavation-induced retaining wall deformation prediction primarily relied on data-driven models trained on field monitoring data from a single excavation project. Zhao et al. (2021) applied a convolutional neural network (CNN) to time–depth matrices constructed from inclinometer measurements collected at one diaphragm wall excavation site. Seo and Chung (2022) evaluated the applicability of 1D-CNN and long short-term memory (LSTM) models using a small dataset compiled from multiple excavation sites, where numerical analysis results were partially incorporated due to limited availability of step-by-step field data. Liu et al. (2023) proposed an empirical mode decomposition (EMD)–recurrent neural network (RNN) framework in which wall deformation time series from a single ultra-deep foundation pit were decomposed into multi-scale components, based solely on field monitoring data. These studies demonstrated the feasibility of learning excavation-induced deformation from monitoring records but were generally constrained by site-specific datasets and limited project diversity.

More recent studies have focused on improving spatiotemporal representation and temporal dependency modeling by adopting hybrid and attention-based deep learning architectures. Gao et al. (2025) developed a convolutional gate recurrent unit (ConvGRU)-based encoder–decoder model with cross-attention, trained on field monitoring data from one excavation project with two inclinometer sections. Liu et al. (2025) and Peng et al. (2025) employed CNN–LSTM-based hybrid models enhanced with attention mechanisms, using multivariate field monitoring data such as wall displacement, ground settlement, groundwater level, and strut axial force, but both relied on single-site datasets. He et al. (2025) introduced a spatiotemporal graph convolutional network (STGCN) to fuse multi-source monitoring data from one subway excavation project, explicitly modeling spatial correlations among monitoring points. Tao et al. (2024) proposed a sequence-to-sequence convolutional long short-term memory (ConvLSTM) framework trained on dense inclinometer data from a single large excavation site, with limited validation on an adjacent excavation zone within the same project. Several studies attempted to enhance generalization or data diversity through alternative data strategies. Le et al. (2025) trained CNN- and LSTM-based models using a mixed database of published field cases and finite element simulation results from multiple projects, although the prediction target was limited to maximum wall deflection rather than full spatiotemporal deformation. Seo and Chung (2025) developed an attention-based LSTM model trained exclusively on time-series inclinometer data collected from three independent excavation sites, explicitly evaluating cross-site applicability. Zhao et al. (2024) proposed a domain-knowledge-guided deep neural network trained on high-resolution field monitoring data from a single site, embedding known deformation mechanisms to mitigate overfitting under limited data conditions.

Despite differences in model architecture and input configurations, existing studies share fundamental characteristics. First, training and validation strategies predominantly rely on the same excavation sites or projects sharing similar excavation contexts and geotechnical conditions, which constrain the assessment of site-independent generalization. Second, most models are trained directly on field monitoring data that inherently includes site-specific bias and measurement noise; as a result, each model learns different error structures and uncertainty characteristics, making consistent cross-model analysis difficult or infeasible (Ying, 2019). These aspects are closely related: when a limited variety and quantity of field measurement is exclusively used for both training and validation, reported prediction performance and characteristics become therefore inherently site-dependent, and objective evaluation of model transferability and uncertainty across different excavation projects becomes difficult.

In contrast to the above approaches, our previous research (Kim and Youn, 2026) proposed a multi-resolution ConvLSTM ensemble framework trained exclusively on numerical analysis data and applied it to two independent excavation sites. The framework avoided information leakage associated with site-specific field measurements and demonstrated that a model trained without direct exposure to field data could achieve a reasonable level of predictive accuracy when applied to real excavation cases. However, the validation was limited to a small number of sites and primarily served to demonstrate feasibility, rather than to provide a systematic or comparative evaluation across diverse excavation conditions.

Motivated by these limitations, the present study adopts the multi-resolution ConvLSTM ensemble framework (Kim and Youn, 2026) and extends its validation to a broader range of excavation sites with varying construction and geotechnical conditions. While the previous study established the feasibility of numerical analysis-driven learning, the present work mainly focuses on systematic site-wise validation. The primary objective is to quantitatively evaluate prediction accuracy and deformation pattern characteristics across multiple sites in a consistent manner. By establishing comparative reference across diverse excavation contexts, this study aims to provide a practical benchmark for assessing the site-independent applicability of the framework, thereby supporting their reliable deployment in excavation safety assessment and construction management.

## 3. Methodology

### 3.1. Multi-Resolution ConvLSTM Framework

Fig. 1 presents the multi-resolution ConvLSTM ensemble framework for predicting retaining wall deformation during staged excavation. The framework integrates ConvLSTM models trained at different temporal resolutions using a stacking ensemble strategy. For database generation, time-series wall deformation profiles were derived from numerical simulations, and data augmentation via Gaussian noise injection was applied. Three ConvLSTM models were developed using excavation histories at different temporal resolutions. Specifically, Model A, Model B, and Model C use 3, 6, and 10 excavation steps, representing deformation responses over excavation depths of 1.5 m, 3.0 m, and 5.0 m, respectively. Each model performs one-step-ahead prediction based on its input sequence and generates multi-step forecasts through a recursive prediction strategy. The resolution-specific multi-step predictions are then combined using a deep-learning-based meta-model, which integrates complementary temporal information and corrects systematic prediction errors. Through this process, the framework produces predictions of retaining wall deformation associated with additional excavation of up to 5.0 m. The detailed network architectures and training configurations follow our previous work (Kim and Youn, 2026).

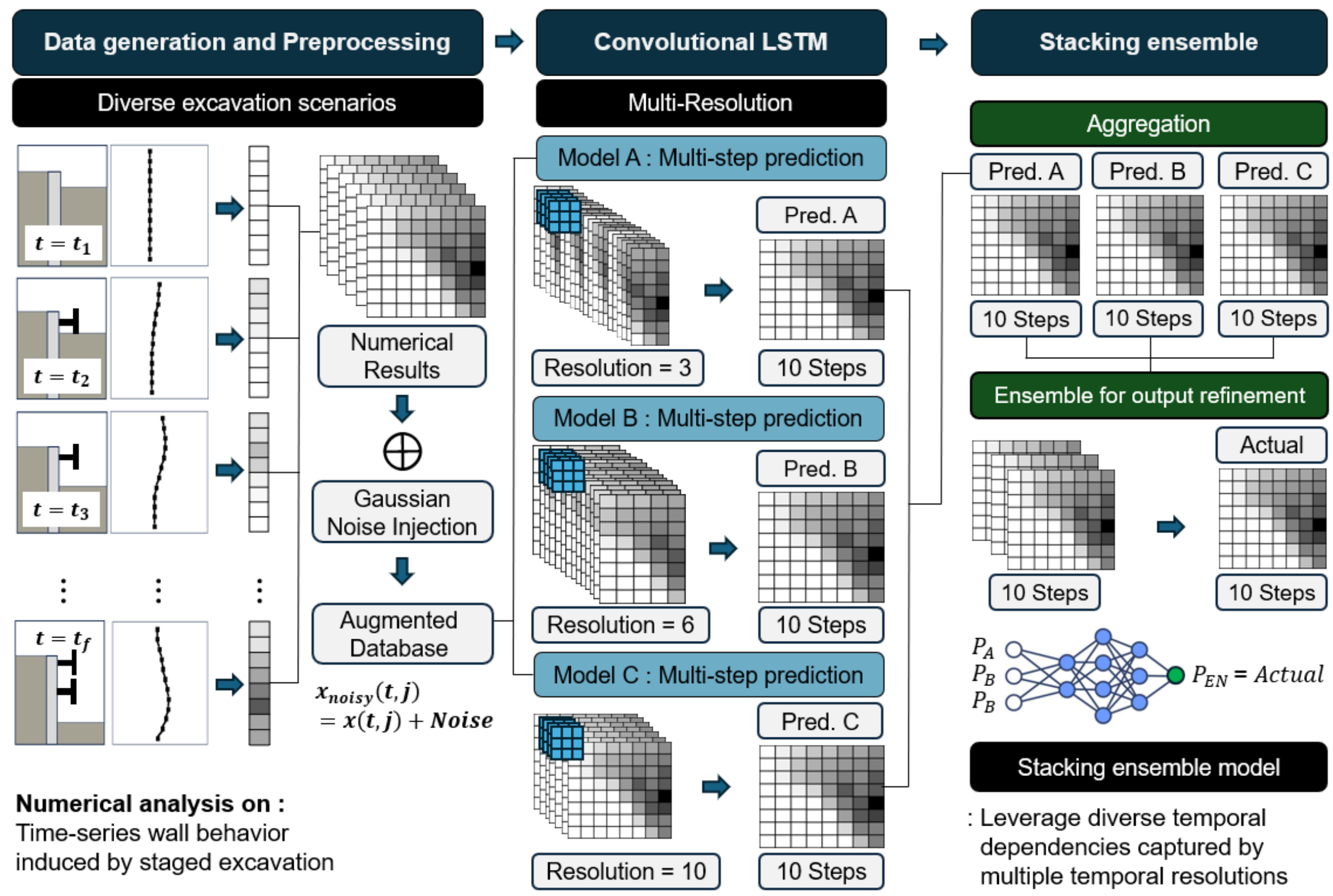

**Fig. 1. Multi-resolution ConvLSTM framework (Modified from Kim and Youn, 2026)**

## 3.2. Data Generation and Augmentation

### 3.2.1. Numerical models and representative results

Fig. 2 illustrates the numerical models used to generate excavation-induced retaining wall deformation data. The models represent typical urban excavation conditions and consist of layered ground profiles, retaining wall systems, and staged excavation sequences. Three excavation cases were defined according to the final excavation depth: Case A (10 m), Case B (14 m), and Case C (20 m), selected to produce distinct wall deformation behaviors associated with different excavation depths and boundary conditions. To account for diversity and uncertainty in ground and structural conditions, key geotechnical and structural parameters were treated as random variables and sampled from predefined probability distributions. The detailed cross-sectional configurations, constitutive models, and statistical definitions of input parameters also follow our previous study (Kim and Youn, 2026).

Excavation was simulated as a sequence of discrete stages with a constant vertical increment of 0.5 m, enabling excavation-induced wall deformation to be captured as time-series responses. For each excavation case, 1,400 numerical simulations were performed. Among them, 1,000 simulations were conducted without groundwater effects, while the remaining 400 simulations incorporated construction-induced groundwater drawdown. In the latter, the initial groundwater level was randomly assigned as 5 m, 10 m, or 15 m below the ground surface, and once excavation reached this level, the groundwater table was assumed to lower following the excavation face in subsequent stages. This modeling scheme allows the influence of groundwater conditions on wall deformation to be systematically reflected.

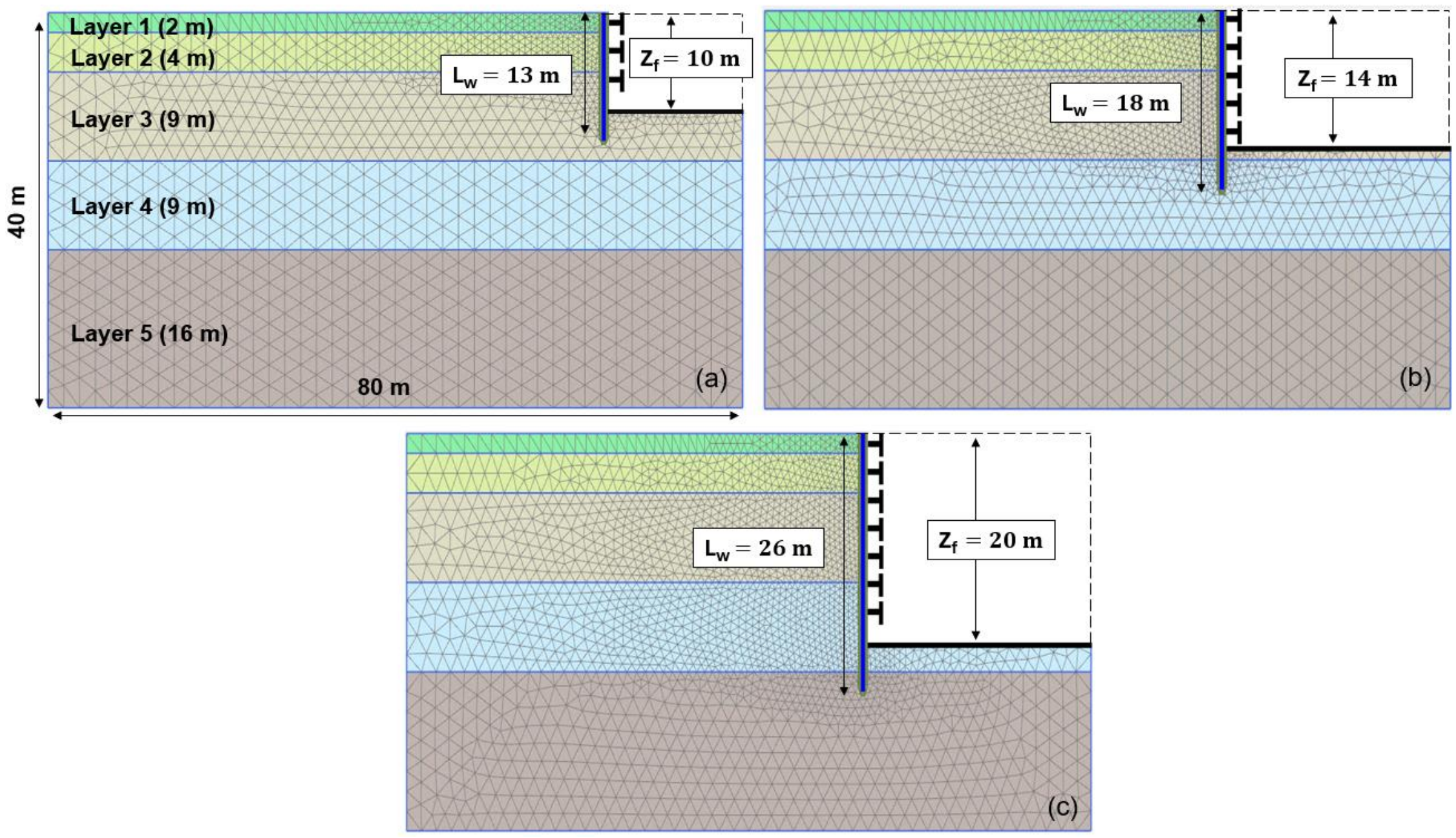

**Fig. 2. Numerical models of excavation using braced retaining wall with excavation depths of: 10 m (Case A), 14 m (Case B), and 20 m (Case C)**

Fig. 3 presents representative time-series wall deformation profiles for Cases A, B, and C for demonstration. In Case A, the retaining wall shows relatively stiff response with small lateral displacements due to shallow excavation, whereas larger deformation magnitudes and curvature are observed in Cases B and C as excavation depth increases. In Case C, wall embedment into the bedrock leads to constrained deformation near the base and more pronounced deformation in the upper portions of the wall.

In each simulation, lateral wall displacements were extracted at monitoring points located at fixed depth intervals along the wall for every excavation stage, forming time-series deformation profiles. To enable consistent learning and prediction for walls with different lengths, the extracted displacement profiles were spatially normalized and resampled to 100 normalized points along the wall length. After preprocessing, the datasets for Cases A, B, and C were organized as four-dimensional tensors of size (1,400, T, 100, 1), where T denotes the number of excavation stages for each case. These datasets were used as the base data for subsequent augmentation and model training.

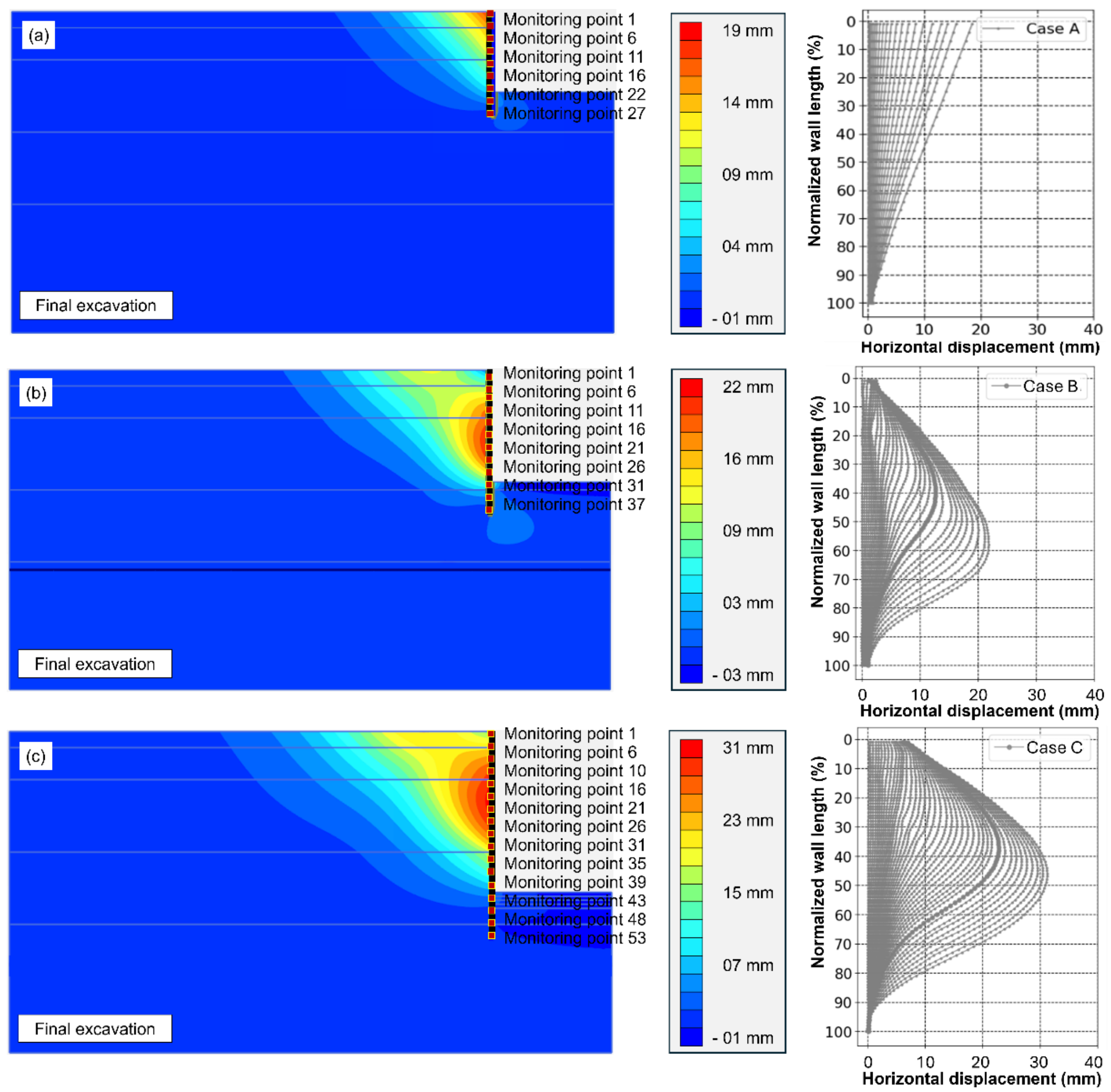

**Fig. 3. Representative numerical results and the wall deformation of (a) Case A, (b) Case B, and (c) Case C**

### 3.2.2. Data augmentation via Gaussian noise injection

While numerical simulations typically generate smooth and idealized deformation curves, field measurements often exhibit spatially irregular wall deformations due to measurement noise, construction-induced discontinuities, and stress redistribution near support elements. To account for this discrepancy and better approximate field-observed behavior, stochastic perturbations based on Gaussian random noise (Shi et al., 2025) were introduced into the numerical displacement data. A one-dimensional Gaussian noise field was first generated along the normalized wall depth:

$$\epsilon_c(i) \sim \mathcal{N}(0,1), i = 1, \dots, R$$

where $R$ denotes the spatial resolution of the noise field. These independent standard normal variables define the spatial structure of the perturbations. The parameter $\sigma$, defined as the standard deviation of the noise field, controls the magnitude of the introduced deformation irregularity.

The augmented displacement field is formulated as

$$x_{\text{noisy}}(t,j) = x(t,j) + \frac{t-1}{t_f-1} \cdot \sigma \cdot \text{Interp}\left(\left\{\frac{i-1}{R-1}, \epsilon_c(i)\right\}_{i=1}^{R}, \frac{j-1}{N-1}\right)$$

where $x(t,j)$ represents the original wall displacement at excavation stage $t$and spatial index $j$, $t_f$ is the final excavation stage, and $N$ (100 in this study) is the number of spatial points along the normalized wall depth. The operator $\text{Interp}(\cdot)$ linearly interpolates the discrete noise field onto the target spatial grid, ensuring spatial continuity of the perturbation. In this formulation, $R$ governs the spatial smoothness of the irregularities: smaller values generate spatially coarse variations, whereas larger values produce more localized fluctuations. The parameter $\sigma$ determines the amplitude of deformation perturbations. The temporal scaling factor $\frac{t-1}{t_f-1}$ progressively increases the perturbation level as excavation advances. This scaling reflects field behavior in which deformation irregularity tends to intensify after support installation. As excavation proceeds and structural supports such as struts or anchors are activated, stress redistribution and local stress concentration frequently occur near support levels, resulting in spatially non-uniform wall deformation. By gradually amplifying the perturbations with excavation stage, the augmentation scheme mimics this increasing irregularity associated with soil–structure interaction and support-induced stress concentration.

In this study, σ was defined as 2%, 6%, and 10% of the maximum displacement from each simulation case. For each selected σ level, a random scaling factor $R$, discretely assigned from 1 to 100 (i.e., 1, 10, 20, 30, …, 100), was applied to introduce variability in the noise amplitude. By applying this augmentation strategy to the original 4,200 numerical results, the database was expanded to a total of 16,800 displacement profiles. Fig. 4 illustrates examples of augmented wall deformation profiles generated using discrete values of σ and $R$.

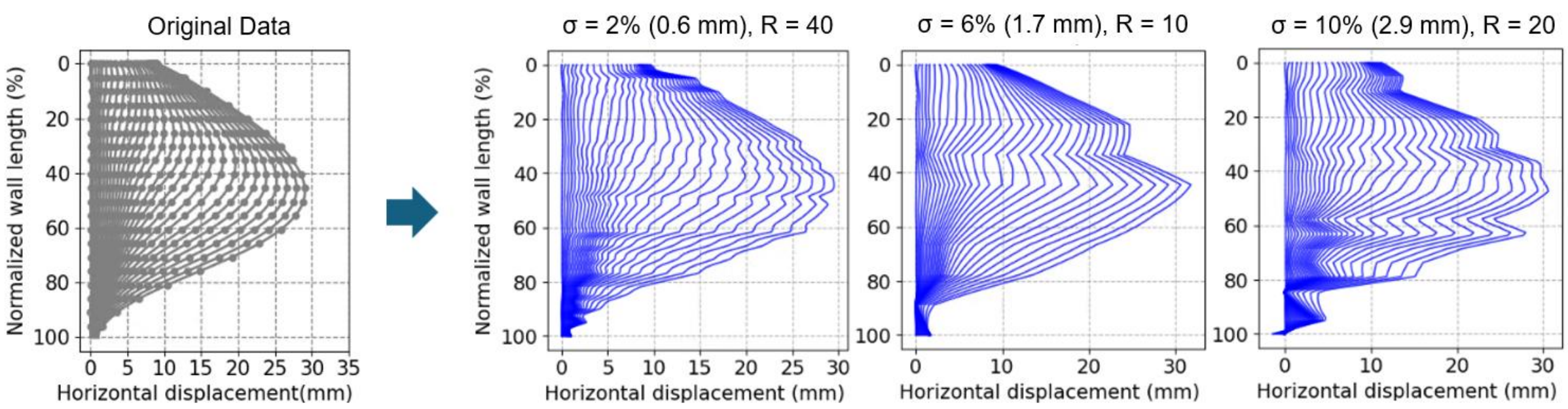

**Fig. 4. Influence of noise parameters on augmented wall deformation**

After data augmentation, the displacement datasets were organized into four-dimensional tensors of size (5,600, 21, 100, 1) for Case A, (5,600, 29, 100, 1) for Case B, and (5,600, 41, 100, 1) for Case C. Multi-step input sequences at different temporal resolutions were generated using a sliding-window approach along the excavation-phase dimension. As a result, the total number of sequences obtained across all cases was 476,000, 425,600, 358,400 for the temporal resolution of 3, 6, and 10, respectively. The generated datasets were randomly divided into training, validation, and test sets using a 7:2:1 ratio.

### 3.3. Field Validation

#### 3.3.1. Excavation sites

Table 1 summarizes the excavation and geotechnical conditions of the 11 excavation sites distributed across South Korea used in this study for field validation. For each site, the number of installed inclinometers, retaining wall and support system type, final excavation depth, and initial groundwater level are presented. The stratigraphy overlying the bedrock is also described with the thickness of each layer, and key site characteristics as well as observed wall deformation patterns are systematically organized. The selected sites encompass a broad range of excavation environments, including variations in excavation depth, groundwater conditions, support configurations (e.g., strut-only systems, combined strut and earth anchor (E/A) systems, and slab supports), and retaining wall types such as cast-in-place (CIP) pile walls, PHC pile walls, diaphragm walls, and H-pile and lagging walls. The initial groundwater level ranges from near-ground-surface conditions at certain sites to considerably deeper levels at others, also indicating differing degrees of groundwater influence during excavation.

When classified in terms of excavation depth, the sites exhibit distinct differences in wall deformation behavior. Relatively deeper excavations (refer to Sites B and D in Table 1) tend to show larger lateral wall displacements accompanied by bulging-type deformation, characterized by the concentration of outward displacement in the mid-depth region of the wall. This deformation pattern reflects a comparatively flexible structural response of the retaining system under sustained excavation-induced loading. In contrast, relatively shallower excavations (refer to Sites G and I in Table 1) generally exhibit smaller displacement magnitudes and a stiffer response, with maximum wall displacement occurring near the top and progressively decreasing toward the base, and with limited development of bulging behavior. On the other hand, embedment conditions also affect the observed wall responses. Sites where retaining walls are embedded into weathered rock or soft rock layers commonly display relatively strong base restraint, whereas walls embedded in comparatively softer soil layers tend to exhibit weaker restraint near the base. In addition, localized undulating deformation patterns are observed at several sites, particularly in cases employing earth anchor systems, suggesting that support configuration also contributes to the detailed shape of the deformation profile.

Overall, Table 1 demonstrates that the validation dataset covers a wide spectrum of excavation depths, groundwater conditions, structural systems, and geotechnical settings. By incorporating multiple sites with markedly different field conditions, the proposed framework is evaluated against diverse real-world scenarios, supporting the robustness and general applicability of the validation results.

Table 1. Summary of excavation and geotechnical conditions of the validation sites

| Site | No. of Inclino-meters | Wall type (support type) | Excavation depth (m) | Initial G.W.L (m) | Soil profile overlaying bedrock (thickness in m) | Site characteristics and observed wall behavior |
|---|---|---|---|---|---|---|
| **A** | 6 | CIP Wall (Strut and E/A) | 14.5-17.5 | 7.1-9.8 | RS(2-3) / WS(7-12) / WR(2-4) | - Embedded in SR; Strong base restraint<br>- Localized undulating wall deformation<br>- Flexural-dominant (bulging-type) wall response |
| **B** | 5 | CIP Wall (Strut) | 16.5-25.0 | 2.4-2.8 | RS(1-3) / AS(3-4) / WS(7-13) / WR(5-) | - Dominant depth-wise distribution of WS/WR<br>- Embedded in WR/SR; Partial base restraint<br>- Flexural-dominant (bulging-type) wall response |
| **C** | 2 | CIP Wall (Strut and E/A) | 10.0-18.0 | 7.7-11.7 | RS(1-3) / AS(2-4) / WS(1-10) / WR(4-) | - Silty clay in the AS<br>- Embedded in SR; Strong base restraint<br>- Flexural-dominant (bulging-type) wall response |
| **D** | 4 | PHC Pile Wall (Strut) | 24.0-26.0 | 17.8-18.8 | RS(3-6) / AS(6-10) / WS(4-18) / WR(5-15) | - Embedded in WS/WR/SR; Partial base restraint<br>- Localized undulating wall deformation<br>- Flexural-dominant (bulging-type) wall response |
| **E** | 3 | Diaphragm Wall (Slab) | 15.0-18.0 | 5.2-5.5 | RS(1-6) / AS(6-10) / WS(7-13) / WR(1-5) | - Embedded in WR/SR; Partial base restraint<br>- Localized undulating wall deformation<br>- Flexural-dominant (bulging-type) wall response |
| **F** | 3 | CIP Wall (Strut) | 15.0-16.0 | 8.1-9.0 | RS(1-3) / AS(9-11) / WS(1-11) / WR(1-3) | - Embedded in WR/SR; Partial base restraint<br>- Localized undulating wall deformation<br>- Flexural-dominant (bulging-type) wall response |
| **G** | 3 | CIP Wall (Strut) | 7.0-8.0 | 9.0-9.7 | N/A | - No groundwater influence during excavation<br>- Shallow excavation depth<br>- Rotation-dominant (cantilever-type) wall response |
| **H** | 2 | CIP Wall (Strut and E/A) | 10.0-11.0 | 4.0-6.3 | RS(2-4) / AS(2-3) / WS(3-6) / WR(3-6) | - Embedded in SR; Strong base restraint<br>- Shallow excavation depth<br>- Flexural-dominant (bulging-type) wall response |
| **I** | 2 | CIP Wall (Strut) | 9.5-10.5 | 2.6-4.1 | RS(5-7) / WS(9-11) / WR(5-) | - Embedded in WS; No base restraint<br>- Shallow excavation depth<br>- Rotation-dominant (cantilever-type) wall response |
| **J** | 2 | CIP Wall (Strut) | 12.5-17.5 | 3.7-4.1 | RS(4-5) / WS(4-6) | - Embedded in SR; Strong base restraint<br>- Flexural-dominant (bulging-type) wall response |
| **K** | 1 | H-Pile Wall (Strut) | 15.0-16.0 | 11.2-11.9 | RS(10-11) / AS(7-8) / WR(1-7) | - Dominant depth-wise distribution of RS/AS<br>- Embedded in RS/AS; No base restraint<br>- Flexural-dominant (bulging-type) wall response |

• RS, AS, WS, WR indicate Reclamation Soil, Alluvial Soil, Weathered Soil, and Weathered Rock layer, respectively.

#### 3.3.2. Representative site descriptions (Sites A and B)

In this section, representative excavation and measurement descriptions of Sites A (Fig. 5) and B (Fig. 6) are respectively illustrated. Descriptions of the other sites (Sites C-K) are included in the Appendix A.

**Site A**

In Site A, the excavation was conducted over an area approximately 80 m in width and 60 m in length. A CIP wall with combined strut and earth anchor supports was employed, and the excavation depth ranged from about 14.5 to 17.5 m. The initial groundwater level was identified at depths of about 10.2–13.5 m below the ground surface, indicating that groundwater influence is limited during the early excavation stages. A total of six inclinometer measurements were obtained, including three corner sections (INC-02, 03, and 06) and three central sections (INC-01, 04, and 05). At each inclinometer, lateral wall displacements at corresponding excavation depths ($Z_e$) were measured six to seven times at excavation intervals of 2–3 m, including the initial readings ($Z_e = 0.0$ m) prior to excavation. After completion of the final excavation stage, the measured maximum horizontal displacements ranged from 20 to 40 mm, with larger displacements observed at sections with greater final excavation depths.

Based on borehole logs obtained in the vicinity of the inclinometer locations, the subsurface profile at Site A comprises reclamation soil, weathered soil, weathered rock, and soft rock layers in order with depth. A notable feature of this site is the presence of soft rock layer at a shallow depth, with thicknesses ranging from 3 to 12 m. For each section, the lower part of the wall is therefore restrained due to the increased stiffness provided by the underlying soft rock layer. In addition, the displacement profiles at sections INC-02, 03, and 04 exhibit an undulating wall behavior, which is attributed to the influence of earth anchor prestressing, leading to localized restraint and redistribution of wall deformation along the wall depth.

**Site B**

Fig. 6 illustrates the excavation conditions and field measurements at another excavation site (Site B). The excavation covered an area approximately 120 m in width and 30 m in length. A cast-in-place pile (CIP) retaining wall was supported by struts, and the maximum excavation depth ranged from 16 to 25 m. The initial groundwater level was shallow, located at about 2.4–2.8 m below the ground surface. Field measurements were collected at five inclinometers, including three corner sections (INC-01, 04, and 05) and two central sections (INC-02 and 03). For each inclinometer, 10–12 measurements were conducted at excavation intervals of 2–3 m, and the maximum horizontal wall displacements ranged from 15 to 45 mm.

The subsurface profile at all sections commonly consists of reclamation soil, alluvial soil, weathered soil, and weathered rock layers in order with depth. Notably, the weathered soil and weathered rock layers are dominantly distributed at depths ranging from 14 to 20 m. A soft rock layer is present at greater depths, and its influence is clearly reflected in the measurements at INC-04 and 05, where wall displacements are restrained near the lower portion of the wall due to restraint from the underlying bedrock.

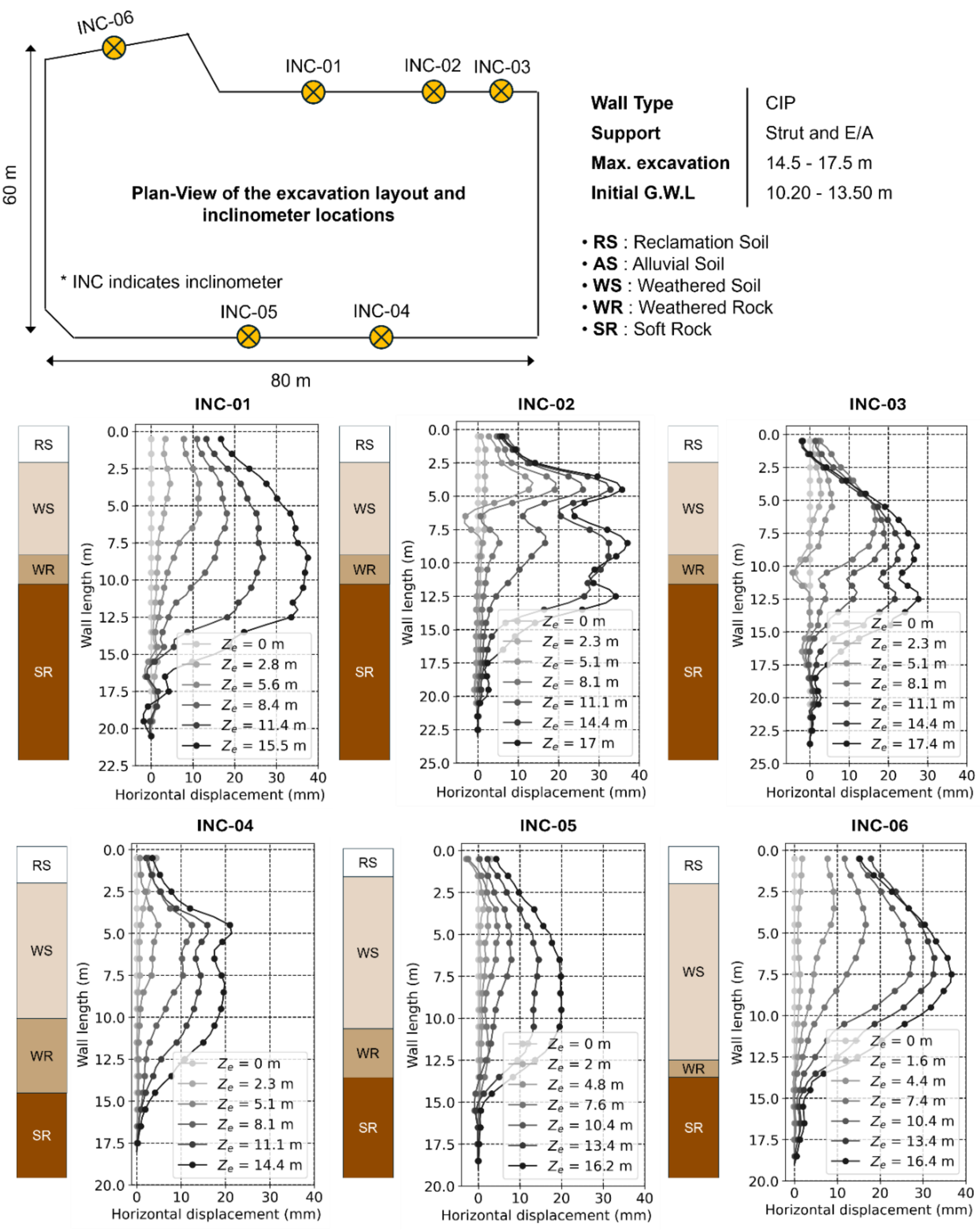

**Fig. 5. Plan view of the excavation layout, inclinometer locations, subsurface stratigraphy, and measured wall displacement profiles (Site A)**

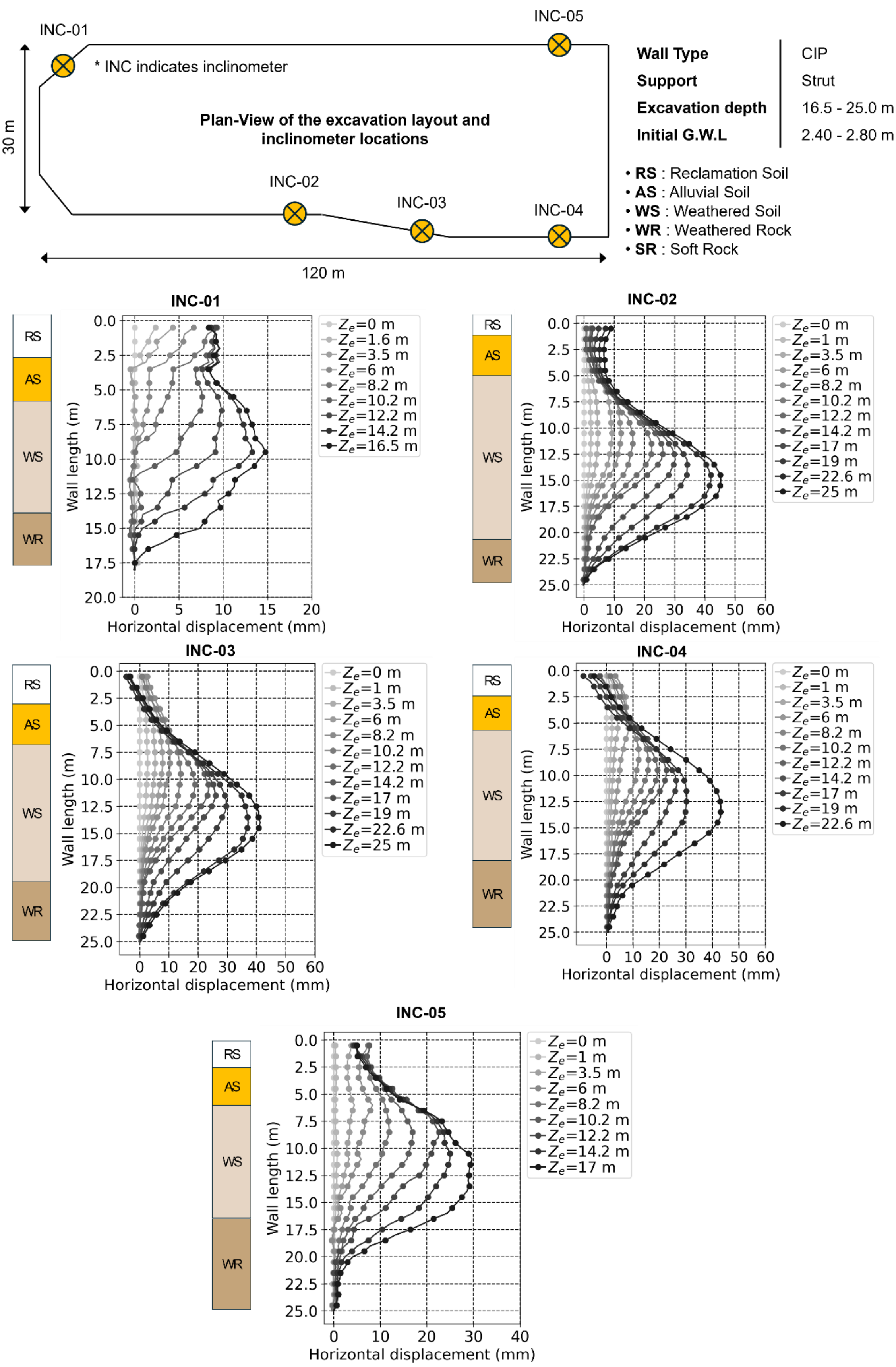

**Fig. 6. Plan view of the excavation layout, inclinometer locations, subsurface stratigraphy, and measured wall displacement profiles (Site B)**

### 3.3.3. Preprocessing field measurements

Fig.7 presents an example of preprocessing and prediction scheme applied to the representative field measurement (INC-01, Site A). The original measurements (Fig. 7(a)) consist of wall displacement profiles recorded at discrete excavation depths ($Z_e$) of 0.0, 2.8, 5.6, 8.4, 11.4, and 15.5 m. Preprocessing is performed in two stages. First, spatial normalization is conducted along the wall length. For each excavation stage, the measured displacement profile is independently interpolated and resampled to uniformly spaced points in terms of normalized wall length, ensuring a consistent spatial resolution (Fig. 7(b)). Second, normalization is carried out in the excavation-depth direction. Because the framework operates at a fixed excavation interval of 0.5 m, the stage-wise profiles are interpolated to generate deformation data at uniform 0.5 m increments, producing a continuous spatiotemporal representation compatible with the model input format (Fig. 7(c)).

Prediction is then performed using the deformation history up to the $N^{\text{th}}$ excavation stage to estimate the wall displacement at the subsequent measured stage ($N+1^{\text{th}}$). Because the framework requires at least 10 previous excavation steps (i.e., a deformation history spanning 5.0 m of excavation with a 0.5 m interval) to generate a prediction, the available target excavation depths for prediction from the dataset are 8.5, 11.5, and 15.5 m (3 samples). Specifically, the wall behavior at an excavation depth of 8.5 m is predicted using deformation measurements accumulated up to 5.5 m, which corresponds to forecasting the wall response over an additional excavation advance of 3.0 m. Similarly, the wall deformation at 11.5 m is predicted based on measurements obtained up to 8.5 m, again representing a forward prediction over an excavation advance of 3.0 m. Finally, the wall displacement at 15.5 m is estimated using deformation data collected up to 11.5 m, which corresponds to predicting the structural response associated with an additional 4.0 m of excavation. The model is then evaluated by comparing the three predicted values ($Z_e$ = 8.5, 11.5, 15.5 m) with their corresponding measurements ($Z_e$ = 8.4, 11.4, 15.5 m), and a detailed description of the evaluation procedure is provided in the subsequent section.

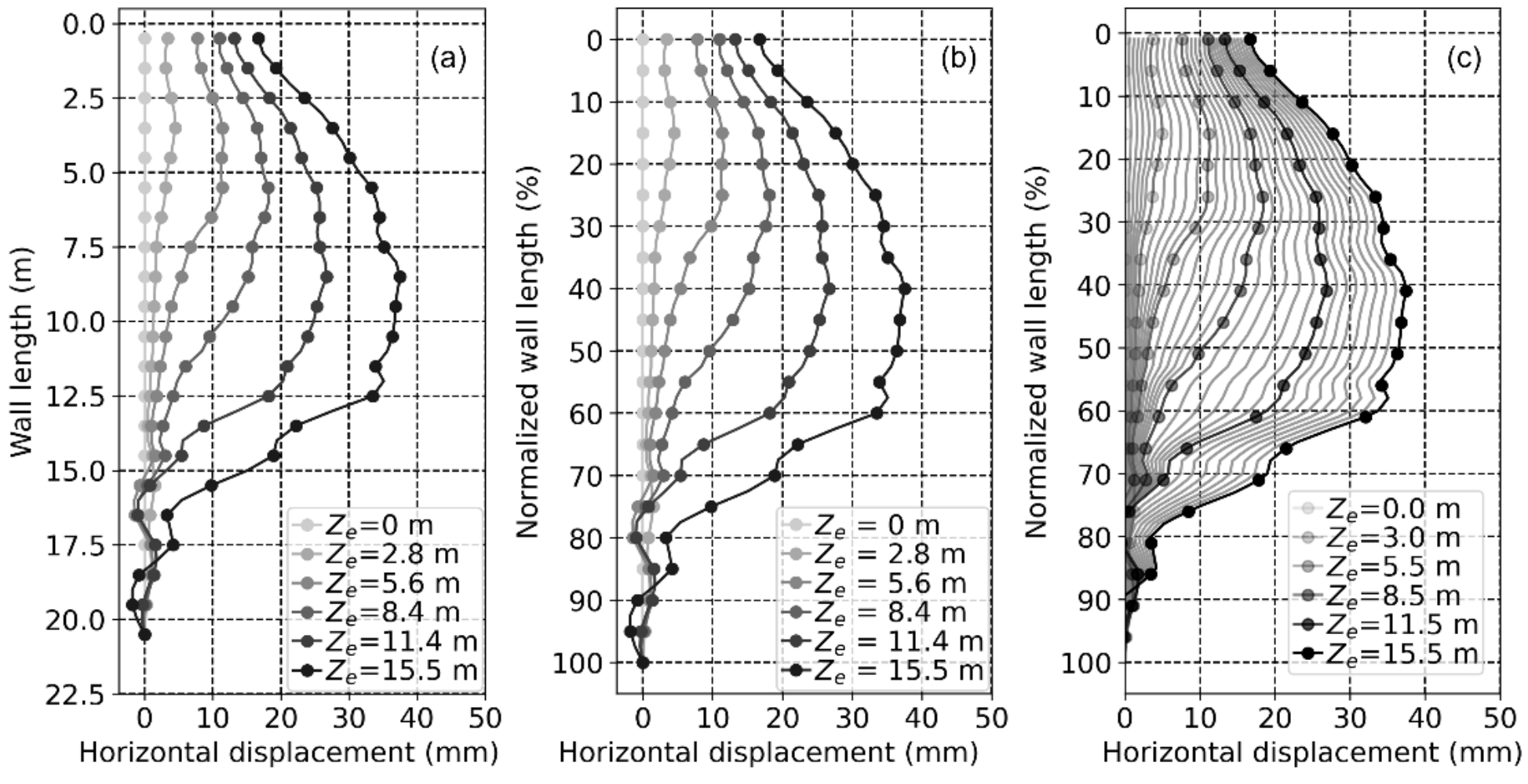

**Fig. 7. Preprocessing procedure: (a) raw, (b) spatially normalized, and (c) temporally interpolated data**

### 3.3.4. Evaluation strategies

The assessment begins with a quantitative comparison between predicted and measured wall displacement profiles at each target excavation stage and progressively extends to detailed spatiotemporal analyses and probabilistic uncertainty quantification.

For a given prediction case, let $y_i$ denote the measured wall displacement and $\hat{y}_i$ the predicted displacement at the $i^{\text{th}}$ spatial location along the wall depth ($i = 1, 2, \ldots, n$), where $n$ is the total number of spatial points. To quantify prediction accuracy, four performance metrics were employed. The mean absolute error (MAE) is defined as

$$\text{MAE} = \frac{1}{n}\sum_{i=1}^{n} | y_i - \hat{y}_i |$$

which represents the average magnitude of prediction error, and the root mean square error (RMSE) is expressed as

$$\text{RMSE} = \sqrt{\frac{1}{n}\sum_{i=1}^{n}(y_i - \hat{y}_i)^2}$$

which penalizes larger deviations more strongly due to the squared term. The coefficient of determination ($R^2$) is defined as

$$R^2 = 1 - \frac{\sum_{i=1}^{n}(y_i - \hat{y}_i)^2}{\sum_{i=1}^{n}(y_i - \bar{y})^2}$$

where $\bar{y}$ denotes the mean of the measured wall displacements, defined as

$$\bar{y} = \frac{1}{n}\sum_{i=1}^{n} y_i$$

and $R^2$ measures the proportion of variance in the measured deformation profile explained by the model. The index of agreement (IoA) (Willmott et al., 2012) is given by

$$\text{IoA} = 1 - \frac{\sum_{i=1}^{n}(y_i - \hat{y}_i)^2}{\sum_{i=1}^{n}(| \hat{y}_i - \bar{y} | + | y_i - \bar{y} |)^2}$$

which provides a normalized measure of agreement between predictions and observations. For each validation site, the performance metrics were computed using all predicted–measured pairs collectively, yielding site-level accuracy indicators.

Beyond global accuracy assessment, representative stepwise predictions were examined to investigate the relationship between prediction performance and the characteristics of the input deformation history. For a target ($N + 1^{\text{th}}$) stage, the model utilizes deformation measurements up to $N^{\text{th}}$ stage to predict the response corresponding to additional excavation depth $\Delta H$ (m). Because the ConvLSTM architecture propagates spatiotemporal features extracted from preceding deformation sequences, the predicted response inherently reflects the deformation trend observed in the immediately preceding stages. Consequently, abrupt variations in deformation increment may result in systematic underestimation or overestimation, revealing the sensitivity of the framework to the spatiotemporal characteristics of the input history.

To further examine spatial and temporal prediction characteristics, depth-wise error distributions were computed along the wall. For each spatial location $i$, the mean prediction error across $m$ validation samples is expressed as

$$\bar{e}_i = \frac{1}{m}\sum_{j=1}^{m}( y_{i,j} - \hat{y}_{i,j})$$

This analysis identifies regions where prediction errors are systematically concentrated, such as mid-depth zones associated with higher deformation demand. In addition, the potential influence of three-dimensional corner effects on prediction accuracy was examined. To this end, prediction performance was evaluated as a function of the normalized wall position ($x/L$), representing the relative distance of the inclinometer from the corner, in order to assess whether proximity to corner regions affects model accuracy. Temporal prediction characteristics were then analyzed by relating performance metrics to the prediction horizon $\Delta H$ (m), defined as the additional excavation depth between the latest available measurement and the target stage. Because the framework performs recursive multi-step prediction up to a maximum horizon of 5.0 m, this analysis quantifies how predictive accuracy degrades as the extrapolation distance increases, reflecting cumulative error propagation over successive stages.

Finally, predictive uncertainty was quantified using Monte Carlo (MC) dropout to enable approximate Bayesian inference. For each input sample, a hundred stochastic forward passes ($T$=100) were performed with different dropout masks. Let $\hat{y}_{i,k}^{(t)}$ denote the predicted displacement at spatial location $i$and prediction step $k$ from the $t$th pass. The predictive mean is computed as

$$\mu_{i,k} = \frac{1}{T}\sum_{t=1}^{T}\hat{y}_{i,k}^{(t)}$$

and predictive uncertainty is defined as the standard deviation:

$$\sigma_{i,k} = \sqrt{\frac{1}{T}\sum_{t=1}^{T}\left(\hat{y}_{i,k}^{(t)} - \mu_{i,k}\right)^2}$$

The resulting uncertainty fields were aggregated across validation samples to construct representative spatiotemporal uncertainty maps, enabling simultaneous characterization of temporal accumulation of uncertainty with increasing prediction horizon and spatial variability along the wall depth. Through this hierarchical evaluation procedure—integrating global accuracy metrics, detailed stage-wise analysis, spatial and temporal performance characterization, and probabilistic uncertainty quantification—the robustness and practical reliability of the proposed prediction framework were comprehensively assessed.

## 4. Results and Discussion

### 4.1. Site-wise Prediction Performance

Table 2 summarizes the site-wise prediction accuracy of the framework, including the number of available inclinometers at each site, together with the corresponding overall MAE, RMSE, IoA, and $R^2$ values. In total, predictions from 85 inclinometers evaluated across 16 validation sites, providing a comprehensive basis for assessing the general performance of the model under diverse excavation and ground conditions. As shown in Table 2, the framework achieved an average MAE of 1.399 mm, an RMSE of 2.144 mm, an IoA of 0.981, and an $R^2$ of 0.930 when all sites were considered together, indicating generally high prediction accuracy and strong agreement with field measurements. Among the individual sites, Site B exhibited the best performance, with MAE = 1.562 mm, RMSE = 2.296 mm, IoA = 0.946, and $R^2$ = 0.984, whereas Site H showed the lowest prediction accuracy, with MAE = 1.419 mm, RMSE = 1.844 mm, IoA = 0.936, and $R^2$ = 0.702.

**Table. 2 Site-wise average prediction accuracy of the wall deformation prediction framework**

| Site | No. of inclinometers | MAE (mm) | RMSE (mm) | IoA | $R^2$ |
|---|---|---|---|---|---|
| **A** | 6 | 2.067 | 2.897 | 0.980 | 0.920 |
| **B** | 5 | 1.562 | 2.296 | 0.984 | 0.946 |
| **C** | 2 | 0.822 | 1.209 | 0.982 | 0.932 |
| **D** | 4 | 0.762 | 1.012 | 0.982 | 0.933 |
| **E** | 3 | 1.586 | 2.617 | 0.964 | 0.829 |
| **F** | 3 | 1.173 | 2.147 | 0.923 | 0.744 |
| **G** | 3 | 0.567 | 0.757 | 0.979 | 0.903 |
| **H** | 2 | 1.419 | 1.844 | 0.936 | 0.702 |
| **I** | 3 | 2.030 | 2.441 | 0.958 | 0.808 |
| **J** | 2 | 0.785 | 1.140 | 0.973 | 0.893 |
| **K** | 1 | 0.752 | 1.059 | 0.981 | 0.923 |
| **All** | 34 | 1.399 | 2.144 | 0.981 | 0.930 |

The scatter plots in Fig. 8 present the relationship between predicted and measured wall displacements for each validation site, together with the corresponding coefficients of determination ($R^2$). For the combined dataset (All Sites), a pronounced linear relationship is observed ($R^2$ = 0.93), indicating that the model explains a substantial proportion of the variance in measured displacements across diverse field conditions. Sites A, B, C, D, G, J, and K ($R^2$ > 0.85) exhibit data points closely aligned with the 1:1 reference line, suggesting consistent predictive performance over the full displacement range. Site B shows the highest agreement ($R^2$ = 0.946), with minimal dispersion around the reference line.

In contrast, Sites E, F, H, and I ($R^2$ < 0.85) display increased dispersion and deviation from the 1:1 line, particularly at moderate to large displacement levels. For Sites F and H ($R^2$ = 0.744 and 0.702, respectively), the scatter becomes more pronounced as displacement increases, indicating reduced predictive accuracy at higher deformation levels. Sites E and I show intermediate behavior, maintaining near-linearity at smaller displacements while exhibiting increasing spread at larger magnitudes.

Overall, among the 11 validation sites, seven sites demonstrate high predictive accuracy ($R^2 > 0.85$), two exhibit moderate accuracy ($0.80 \leq R^2 < 0.85$), and two show comparatively lower accuracy ($R^2 \approx 0.70$). These results indicate that the proposed framework achieves reliable performance in the majority of field cases, with performance degradation confined to a limited number of sites characterized by increased prediction–measurement dispersion. Although the scatter plots demonstrate overall site-level agreement, a more detailed analysis is necessary to clarify the sources of performance variation. The following section therefore investigates spatial error distribution and the influence of prediction horizon.

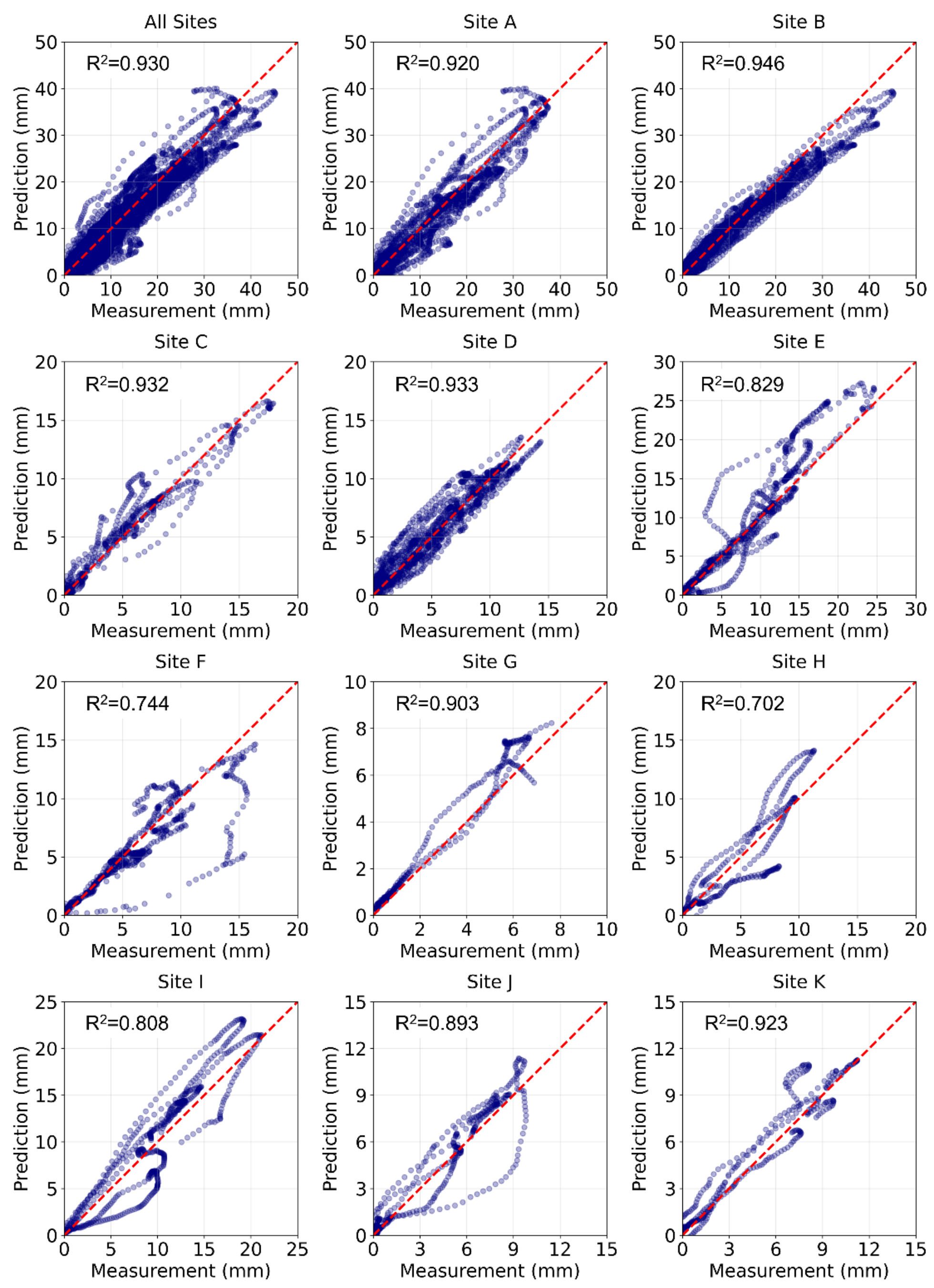

**Fig. 8. Comparison between predicted and measured wall displacements across excavation sites**

### 4.2. Excavation stage-dependent Prediction Performance

To elucidate how the proposed framework translates input deformation characteristics into predictive responses, representative stage-wise analyses were conducted for INC-01 and INC-02 of Site A. The objective is not merely to report prediction accuracy, but to interpret how the model behavior reflects the deformation patterns embedded in the immediately preceding input sequences.

Fig. 9 presents representative stage-wise predictions for INC-01 of Site A by comparing the predicted wall displacement profiles with the corresponding field measurements at each excavation depth ($Z_e$). For INC-01, the predicted profiles exhibit very close agreement with the measured responses across successive excavation stages. At an excavation depth of 8.4 m (Fig. 9(b)), corresponding to approximately 3.0 m of additional excavation beyond the latest monitored depth of 5.5 m, the model achieves high accuracy (MAE = 1.87 mm, RMSE = 2.44 mm, IoA = 0.98, and $R^2$ = 0.96). A similarly high level of agreement is observed at 11.4 m excavation depth (Fig. 9(c)), predicted using measurements up to 8.5 m, with MAE = 1.98 mm, RMSE = 2.61 mm, IoA = 0.98, and $R^2$ = 0.95.

Importantly, the model accurately captures not only the overall deformation magnitude but also the stage-dependent migration of the peak displacement location. The position of maximum displacement shifts progressively with excavation depth—approximately 25% of the normalized wall depth at 8.4 m, about 40% at 11.4 m, and nearly 50% at 15.5 m—and this movement is consistently reproduced in the predictions. In addition, the curvature of the deformation profile and the restrained displacement behavior near the lower portion of the wall are precisely represented. The predicted curves remain nearly coincident with the measured profiles over the full depth range, demonstrating high-fidelity reconstruction of both deformation magnitude and spatial configuration. These results confirm that the spatiotemporal features extracted from the preceding deformation history effectively encode the evolving deformation pattern governing subsequent wall response.

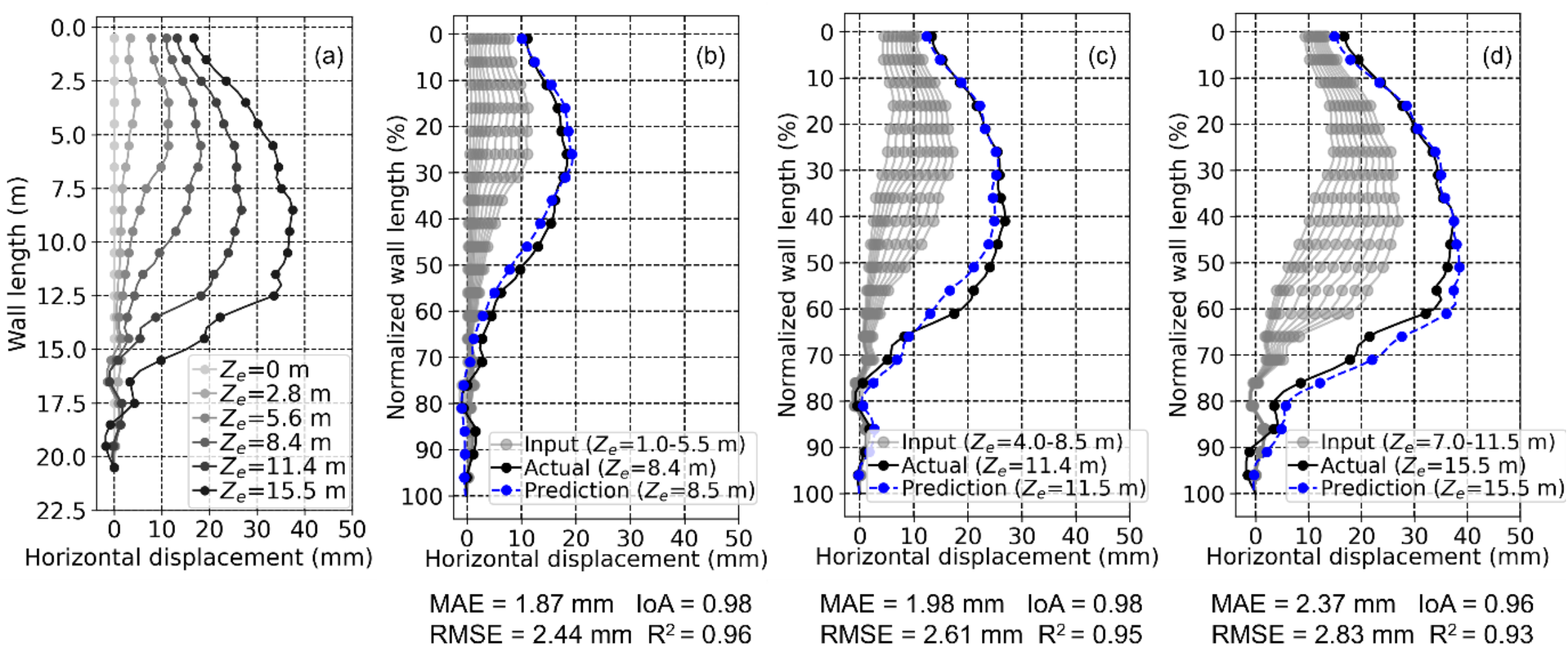

**Fig. 9. Wall deformation prediction for INC-01 of Site A: (a) raw data, predictions at excavation depths of (b) 8.4 m, (c) 11.4 m, and (d) 15.5 m**

While INC-01 illustrates a case in which gradual deformation evolution is translated into highly consistent predictions, INC-02 of Site A (Fig. 10) more clearly reveals the model's dependence on the characteristics of the input deformation history. Although successive excavation intervals correspond to approximately 2.5–3.0 m of additional excavation, the associated deformation increments differ substantially (Fig. 10(a)). Because the model extrapolates from the deformation trend embedded in the most recent input sequence, it implicitly assumes an increment comparable to that observed in the preceding stage. As a result, when the actual increment increases markedly, the predicted displacement is underestimated, whereas a subsequent reduction in increment leads to overestimation. These behaviors indicate that the prediction output is governed primarily by the incremental deformation patterns contained in the input history rather than by excavation depth alone.

This deformation-history dependency is manifested in a spatially non-uniform manner along the wall depth, as shown in Fig. 10(d) and (e). In the upper portion of the wall (approximately 0–20% normalized depth), both measured and predicted displacements remain small, and the model predicts negligible additional movement where the input sequences exhibit minimal variation. In contrast, the mid-depth region (approximately 35–55%) shows pronounced curvature and larger stage-to-stage increments. In Fig. 10(d), the increased actual increment results in slight underestimation near the peak zone, whereas in Fig. 10(e), moderate growth produces localized overestimation. Despite these magnitude differences, the model preserves the overall deformation shape and accurately captures the depth of maximum displacement. In the lower portion of the wall (approximately 60–100%), where structural restraint limits movement, both measured and predicted displacements converge toward near-zero values, and the restrained behavior is consistently reproduced.

Overall, the framework does not impose uniform scaling across the profile; instead, it adjusts predicted magnitudes according to localized deformation characteristics embedded in the prior stages. This reflects ConvLSTM architecture's capacity to propagate sequential spatial features while maintaining depth-dependent deformation patterns. Although localized discrepancies appear in the mid-depth region, the global deformation shape, peak migration, and lower-boundary restraint are consistently reproduced, and the overall prediction performance remains high ($R^2 \geq 0.83$), confirming strong practical reliability.

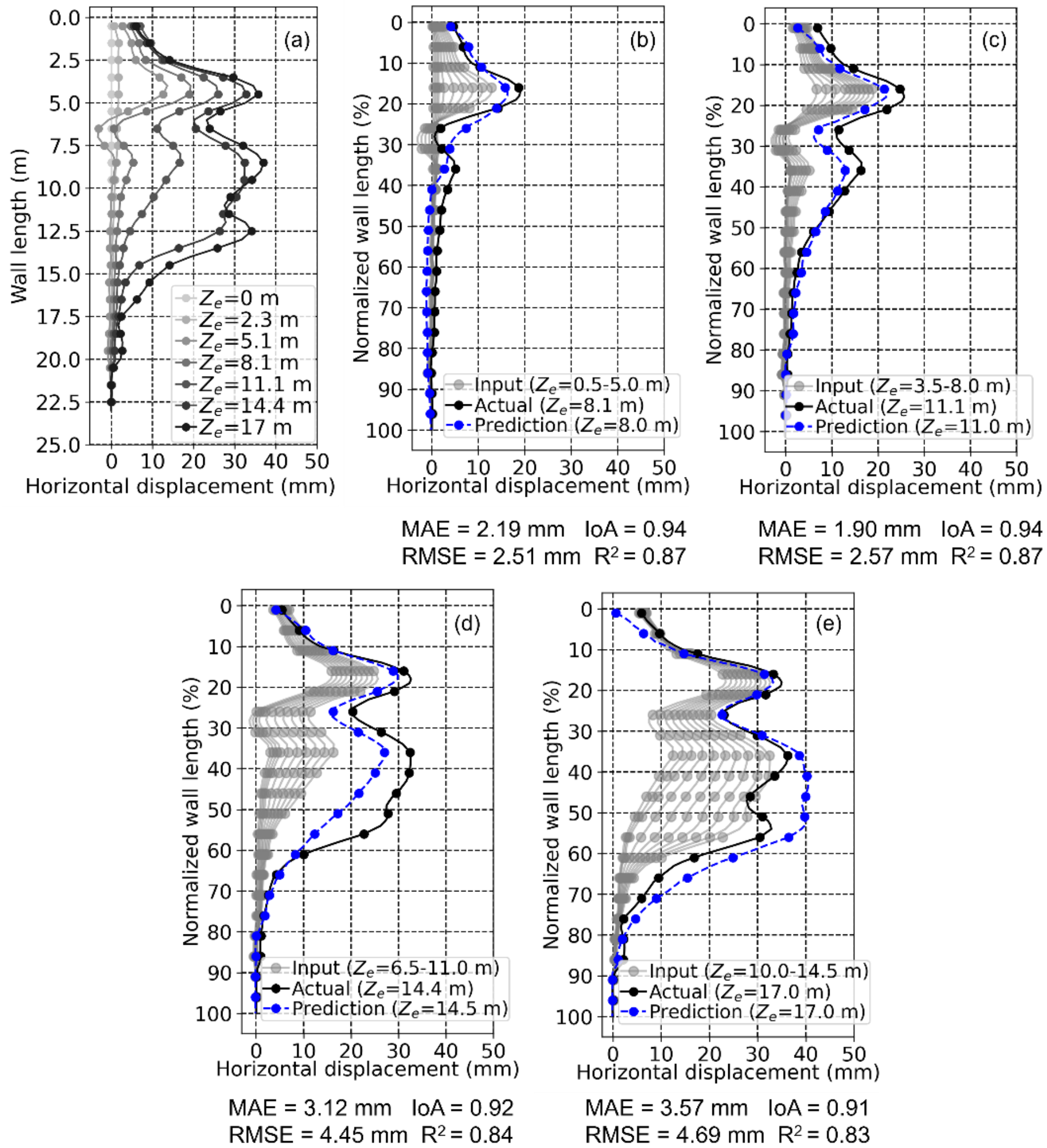

**Fig. 10. Wall deformation prediction for INC-02 of Site B: (a) raw data, predictions at excavation depths of (b) 8.1 m, (c) 11.1 m, (d) 14.4 m, and (e) 17.0 m**

## 4.3. Spatial Prediction Characteristics

### 4.3.1 Depth-wise error distribution

To investigate the spatial characteristics of prediction performance, a depth-wise error analysis was conducted along the normalized wall length, as presented in Fig. 11. Fig. 11(a) shows the distribution of prediction errors for all samples together with the mean and ±1σ (standard deviation) range. Fig. 11(b) illustrates the depth-wise variation of the mean error, while Fig. 11(c) presents the corresponding depth-wise RMSE profile.

As observed in Fig. 11(a), prediction errors are generally small near the top and bottom of the wall, indicating relatively stable predictive performance in these boundary regions. In contrast, larger and more scattered errors appear in the mid-depth zone, approximately between 50% and 80% of the normalized wall length, where overpredictions of up to about 13 mm and underpredictions of up to about 10 mm are observed. This trend is reflected in Fig. 11(b), where a modest systematic underestimation (maximum mean error ≈ −0.75 mm) is identified near the wall midspan, along with a slight overestimation (approximately +0.52 mm) in the lower-mid region. Overall, the mean error remains predominantly negative, indicating a mild global tendency toward underprediction.

The RMSE profile in Fig. 11(c) reveals a clear spatial variation in prediction accuracy. The RMSE increases from the upper region toward the mid-depth zone, reaching a maximum of approximately 2.88 mm at 55–60% of the normalized wall length, before decreasing toward the bottom where it reaches a minimum of about 0.28 mm. Notably, RMSE values are substantially larger than the corresponding mean errors at most depths, indicating that the elevated mid-depth error is dominated by increased dispersion of prediction errors rather than systematic bias.

From a geotechnical perspective, the mid-depth region is commonly associated with the location of maximum lateral displacement and significant curvature change of the retaining wall. In this zone, soil–structure interaction effects are more pronounced, and the deformation response becomes highly sensitive to variations in ground conditions, support stiffness, and excavation stages. As a result, displacement profiles in this region tend to exhibit greater case-to-case variability. This elevated variability makes it more difficult for the model to consistently capture the deformation pattern, leading to increased RMSE values. In contrast, the upper and lower boundary regions are more strongly constrained by boundary conditions, resulting in relatively stable deformation patterns and lower prediction errors. Overall, these results suggest that model performance varies systematically along the wall depth and is closely linked to the underlying mechanical response characteristics.

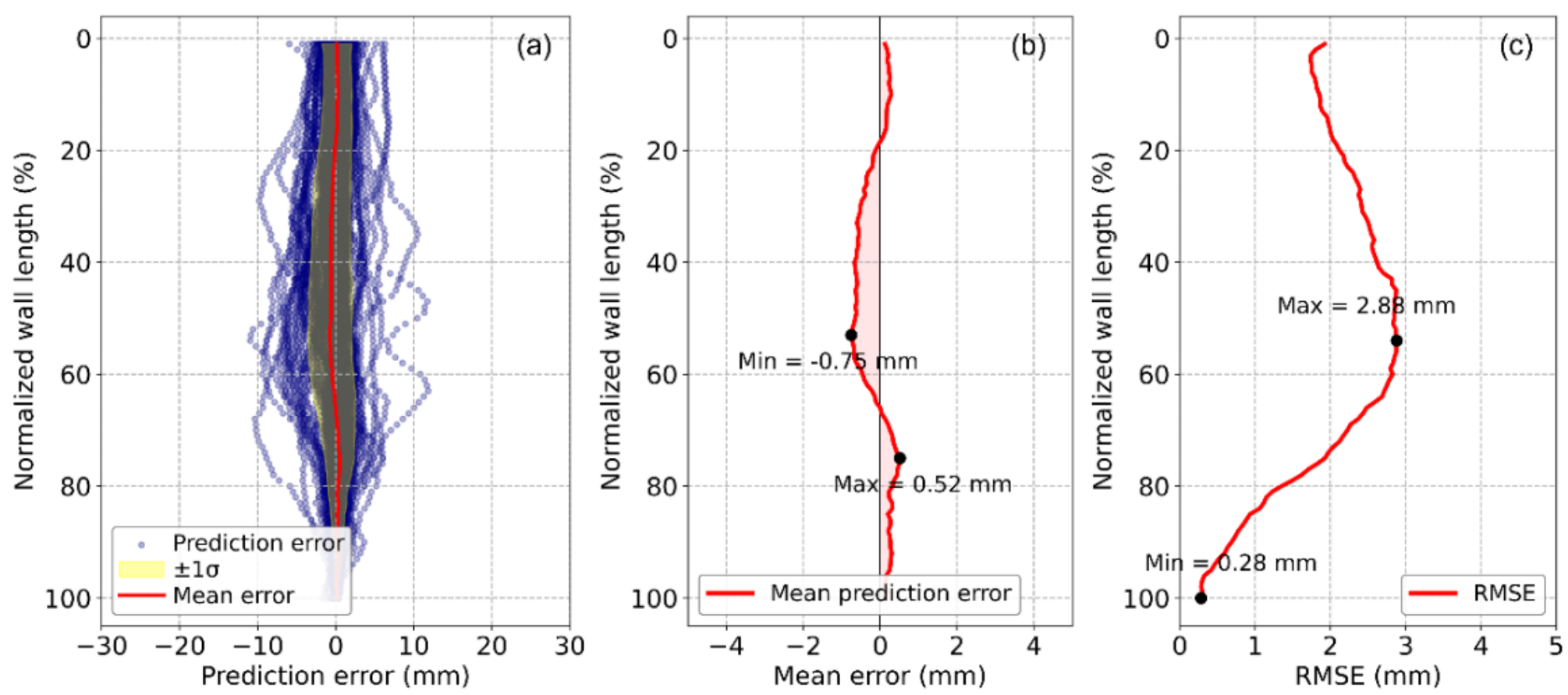

**Fig. 11. Spatial prediction characteristics: (a) prediction error distribution, (b) mean prediction error, and (c) RMSE distribution along the normalized wall length (%)**

**4.3.2 Corner effect**

Retaining walls exhibit inherently three-dimensional behavior, in which stress conditions and deformation patterns vary along the wall depending on spatial position and boundary constraints. In particular, wall segments near corner regions are subjected to three-dimensional restraint due to the interaction with adjacent walls and soils, whereas segments toward the midspan experience relatively less confinement. As a result, larger lateral deformations tend to develop in the midspan region under comparable excavation conditions, reflecting the well-known corner effect (Ou and Shiau, 1998).

To quantify the relative wall location, a corner index (x/L, refer to Fig. 12(a)) was introduced, where smaller values indicate wall segments closer to the corner region and larger values correspond to locations nearer to the wall midspan. As shown in Fig. 12(b), the prediction accuracy, expressed in terms of $R^2$, exhibits a moderate negative correlation with the corner index ($R^2 \approx 0.39$). This trend indicates that prediction accuracy gradually decreases as the wall location shifts from the corner toward midspan. The observed reduction in $R^2$ is consistent with the structural behavior described above: because larger deformations occur toward the midspan due to reduced three-dimensional restraint, the model encounters relatively greater difficulty in reproducing the full variability of displacement patterns in this region. The RMSE distribution shown in Fig. 12(c) provides complementary insight into the absolute magnitude of prediction errors. In contrast to the decreasing $R^2$, the median RMSE shows a gradual increasing trend with the corner index (trendline $R^2 \approx 0.30$). This indicates that as the wall location moves toward the midspan, not only does the model's ability to reproduce detailed deformation patterns decrease slightly, but the absolute prediction error also increases moderately. However, even at the largest corner index values, the median RMSE remains within a practically acceptable range (approximately 2 mm), suggesting that the increase in absolute error is controlled rather than excessive.

Taken together, these results indicate that although prediction performance deteriorates gradually from corner to midspan due to the inherent three-dimensional deformation characteristics of retaining walls, the framework maintains stable and practically reliable predictive capability across different wall locations.

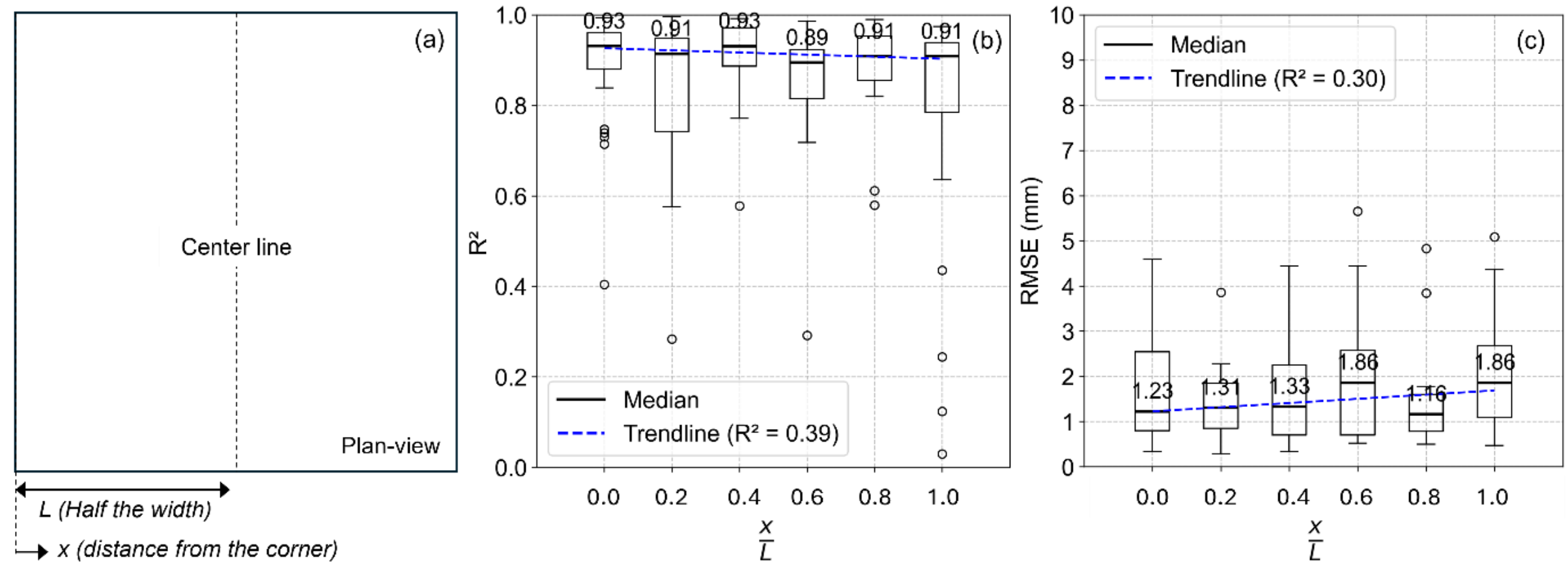

**Fig. 12. Influence of corner effect on predictive accuracy: (a) definition of normalized inclinometer position (x/L), variation of (b) $R^2$, and (c) RMSE with normalized inclinometer position**

### 4.4. Temporal Prediction Characteristics

The validation monitoring data used in this study were collected at different excavation sites, resulting in non-uniform monitoring intervals. Consequently, the excavation increments between the measurements vary across samples, ranging from approximately 1.0 to 5.0 m. To examine the influence of prediction horizon on prediction performance, Fig. 13 presents the variation of both $R^2$ and RMSE with respect to the excavation increment. As shown in Fig. 13(a), the median $R^2$ decreases as the prediction horizon expands. For near-future predictions corresponding to an additional excavation of approximately 1.0 m, the median $R^2$ approaches 0.99. As the horizon increases to 2.0 m, the median $R^2$ remains high (approximately 0.94), but further declines to around 0.60 at a 5.0 m excavation increment. A trend analysis indicates a moderate negative relationship between $R^2$ and prediction horizon ($R^2$ = 0.48), suggesting that the ability of the model to reproduce detailed deformation patterns gradually diminishes with increasing prediction length.

In contrast, the RMSE results shown in Fig. 13(b) reveal a more gradual increase in absolute error magnitude. The median RMSE increases from approximately 0.50 mm at a 1.0 m prediction horizon to about 3.01 mm at 5.0 m. The trendline analysis ($R^2$ = 0.29) indicates a positive but relatively weak relationship between RMSE and prediction horizon. This suggests that while prediction errors accumulate over longer horizons, the growth in absolute error magnitude remains moderate and does not escalate abruptly. The combined analysis of $R^2$ and RMSE indicates that although pattern fidelity decreases with increasing horizon, the absolute error magnitude remains within a practically manageable range for short to intermediate prediction horizons (up to approximately 4.0 m). These results support the applicability of the framework to practical excavation-stage deformation prediction while highlighting the expected degradation associated with extended forecasting horizon.

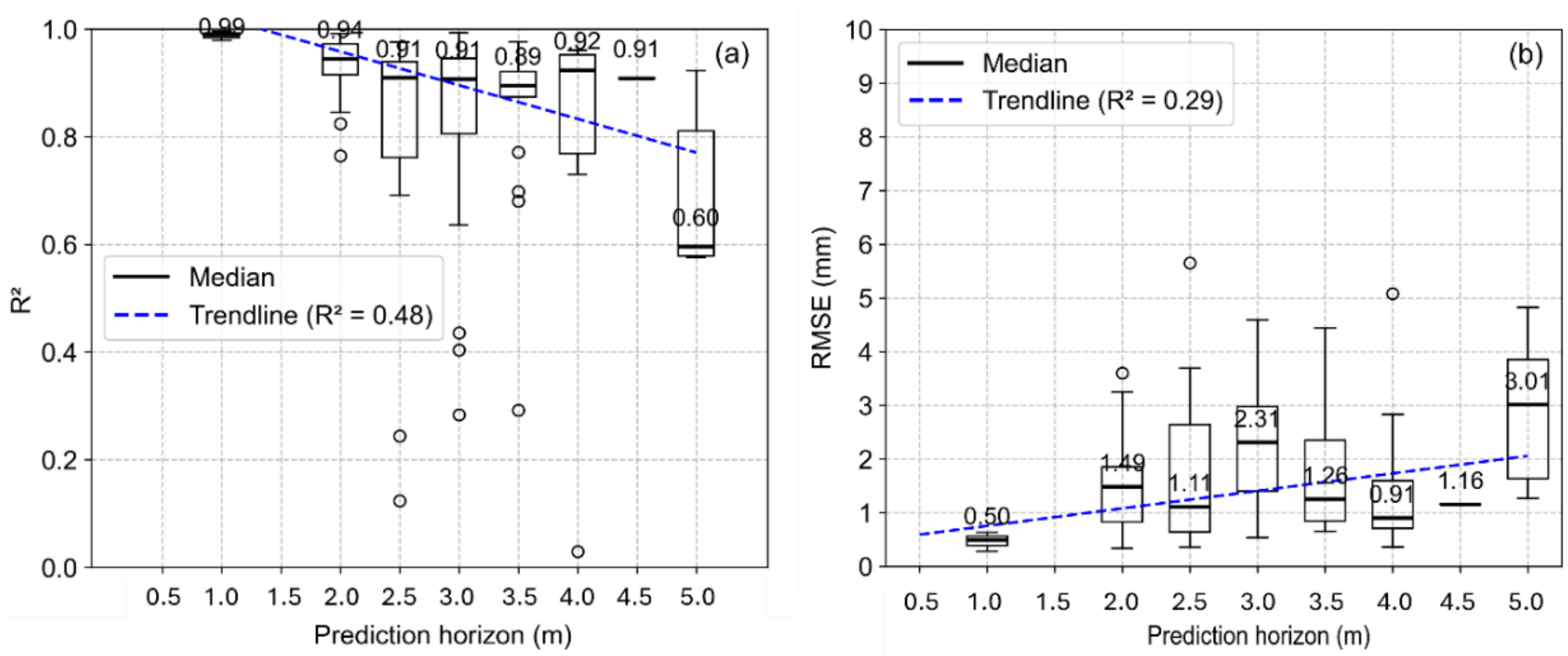

**Fig. 13. Influence of prediction horizon on predictive performance: (a) variation of $R^2$ and (b) variation of RMSE with prediction horizon**

In addition to the overall accuracy metrics, the prediction performance of the peak lateral displacement — a key indicator in construction management — was further evaluated and visualized in Fig. 14. For each prediction step, the maximum absolute displacement was extracted from both the predicted and measured profiles, and the peak displacement error was defined as the absolute difference between these values. The results show that for short prediction horizons (1–2 m of additional excavation), the median peak error remains below approximately 1 mm, indicating highly stable near-term forecasting performance. As the prediction horizon increases, both the median and dispersion of peak error gradually increase. However, even at the longest prediction horizon considered (5 m), the median peak error remains around 3 mm, and no abrupt error escalation is observed aside from a few outliers.

These findings indicate that the magnitude of the structurally critical peak displacement is reproduced with relatively stable accuracy. The quantified distribution of peak displacement error can therefore serve as a practical reference margin for conservative engineering decision-making, supporting the applicability of the framework in excavation-stage monitoring and management.

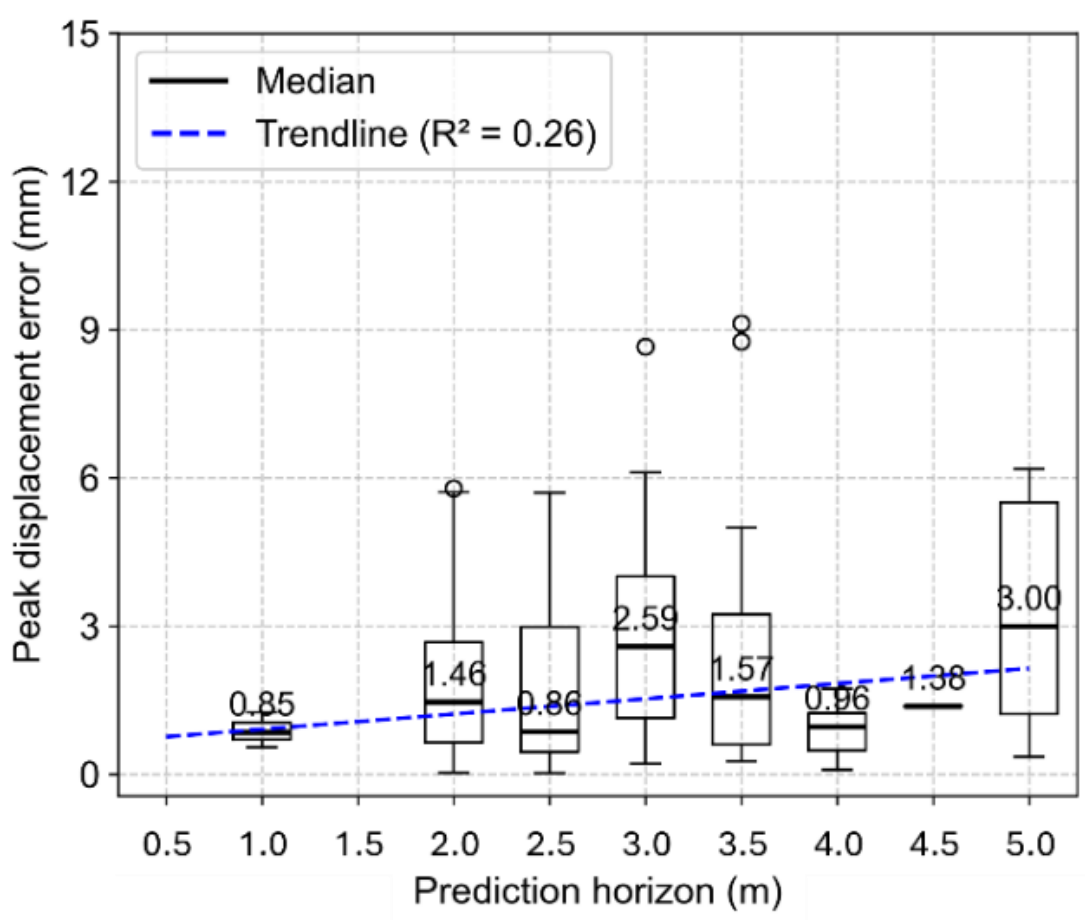

**Fig. 14. Peak displacement error distribution across prediction horizon**

#### 4.5. Spatiotemporal Prediction Uncertainty

Following the analyses of spatial and temporal prediction errors presented in the preceding sections, the predictive uncertainty of the framework is quantified by MC dropout (Milanés-Hermosilla et al., 2021), which enables approximate Bayesian inference by activating dropout layers during inference. For a given input sample, the deep meta model is evaluated repeatedly with stochastic dropout masks, yielding a distribution of predicted wall deformation profiles. The predictive uncertainty at each spatiotemporal point is then defined as the standard deviation of the resulting predictive distribution, which directly reflects the variability of the model outputs under identical input conditions. Specifically, the MC dropout procedure is performed with a hundred stochastic forward passes for each prediction sample. This number was found to be sufficient to ensure stable estimation of the predictive standard deviation while maintaining reasonable computational efficiency. Uncertainty is computed independently at each prediction step and spatial position along the wall, thereby preserving the full spatiotemporal structure of the predictions.

To obtain representative uncertainty characteristics, the above procedure is applied to all available samples. The spatiotemporal uncertainty maps derived from individual samples are then averaged across the dataset, resulting in an aggregated uncertainty field that captures the typical temporal evolution and spatial distribution of predictive uncertainty. The uncertainty values are retained in absolute units (mm) to preserve their physical interpretability, which is directly relevant for engineering assessment and decision-making.

Fig. 15 presents the resulting spatiotemporal distribution of average predictive uncertainty across all validation samples. As shown in the figure, the predictive uncertainty generally increases with prediction horizon, indicating a progressive accumulation of uncertainty as the prediction horizon extends further beyond the current excavation stage. This trend reflects the increasing difficulty of accurately extrapolating wall deformation behavior over larger additional excavation depths. In addition to the temporal effect, a clear spatial pattern is observed along the wall height. Higher uncertainty is consistently concentrated around the mid-height region of the wall, whereas relatively lower uncertainty is observed near the top and bottom boundaries.

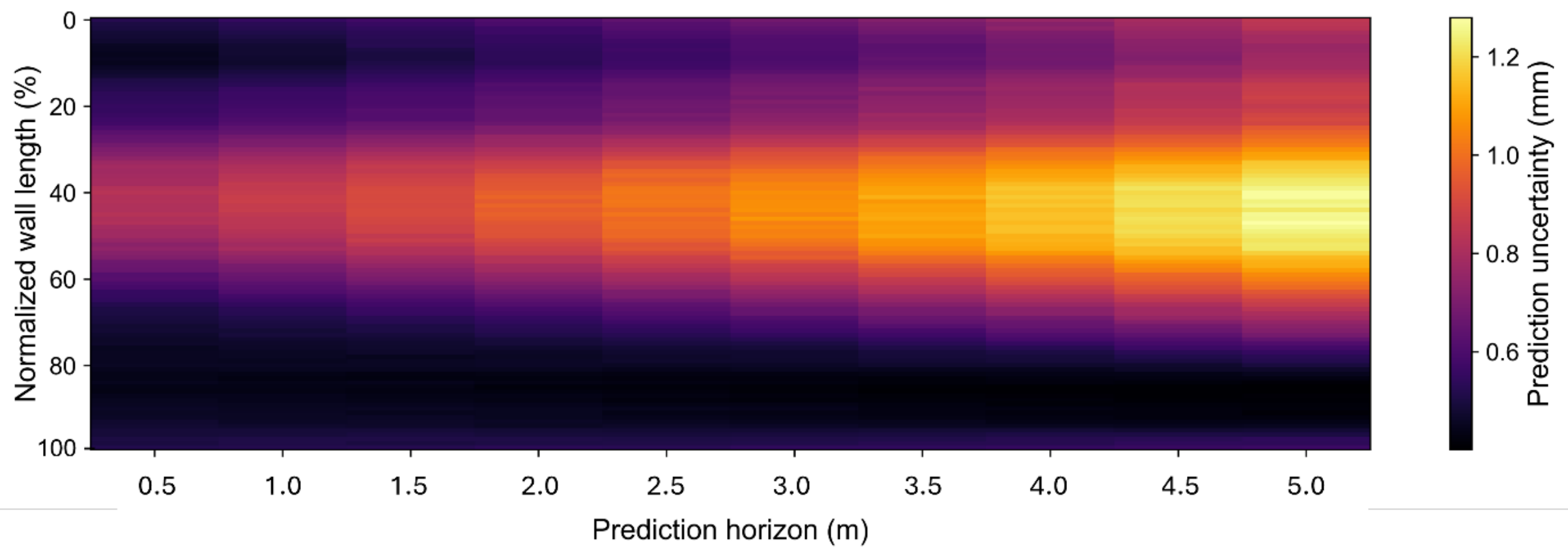

**Fig. 15. Spatiotemporal distribution of average prediction uncertainty across validation sites**

Overall, the results presented in Fig. 15 demonstrate that the framework provides physically consistent and interpretable uncertainty estimates, capturing both the temporal accumulation of uncertainty with increasing prediction horizon and the spatial variability associated with wall deformation characteristics. Importantly, the relatively small magnitude of the estimated uncertainty, compared to typical wall deformation levels observed in excavation projects, suggests that the predictions are not only accurate but also accompanied by a quantitatively reasonable level of uncertainty. These uncertainty patterns complement the prediction accuracy analysis and provide valuable insight into when and where the model predictions should be interpreted with greater caution in practical excavation monitoring and decision-making contexts.

## 5. Conclusion

The field applicability of a multi-resolution ConvLSTM framework was evaluated using 34 inclinometer datasets collected from 11 excavation sites in South Korea. To ensure a site-independent training dataset, comprehensive numerical simulations were performed with stochastically distributed geotechnical and structural properties, and the resulting dataset was expanded to a total of 16,000 samples through Gaussian noise injection. By integrating models of varying temporal resolutions via a stacking ensemble strategy, the framework's spatiotemporal prediction performance was systematically evaluated. The key findings are as follows:

1) The framework provides reliable retaining wall deformation predictions and demonstrates strong field applicability, despite being trained exclusively on numerically simulated and augmented datasets.
2) Across 11 excavation sites, the framework predicts retaining wall deformation associated with up to 5.0 m of additional excavation with a high average accuracy ($R^2$ = 0.93), consistently capturing both the magnitude and spatiotemporal evolution of wall deformation under diverse field conditions.

3) Spatial error analysis along the normalized wall length indicates a systematic yet limited bias, characterized by slight underestimation in the mid-depth region (MAE ≈ 0.75 mm) and relatively small overprediction near the wall top and bottom, confirming stable depth-wise prediction behavior.
4) Prediction performance shows a moderate correlation with the corner index, indicating a gradual decrease in accuracy toward the wall midspan. Nevertheless, prediction accuracy remains high, with $R^2$ of 0.91 even at the midspan. In contrast, temporal prediction accuracy decreases with increasing prediction horizon, exhibiting a moderate negative correlation due to recursive error accumulation.
5) Monte Carlo dropout–based uncertainty quantification reveals very low predictive uncertainty, with a maximum mean value of approximately 1.3 mm, which increases with prediction horizon and is spatially concentrated in the wall mid-depth region, reflecting structurally critical deformation zones.

Despite its high predictive accuracy, the framework's performance was sensitive to temporal irregularities in wall behavior. This limitation suggests that while synthetic datasets capture fundamental geotechnical mechanisms, they may not fully encompass the abrupt, non-linear deformation shifts. However, this discrepancy can serve as a reliable diagnostic indicator; by accurately defining 'expected' behavior, the framework enables engineers to identify anomalous deformations that deviate from standard geotechnical patterns.

**Acknowledgement**

This work was supported by the National Research Foundation of Korea (NRF) grant funded by the Korea government (MSIT) (No. 2023R1A2C1007635).

**Data Availability**

The data that support the findings of this study are available from the corresponding author, H. Youn, upon reasonable request.

**Declarations**

No potential conflict of interest was reported by the authors.

**Appendix A. Plan view of the excavation layout, inclinometer locations, subsurface stratigraphy, and measured wall displacement profiles for excavation sites (C-K)**

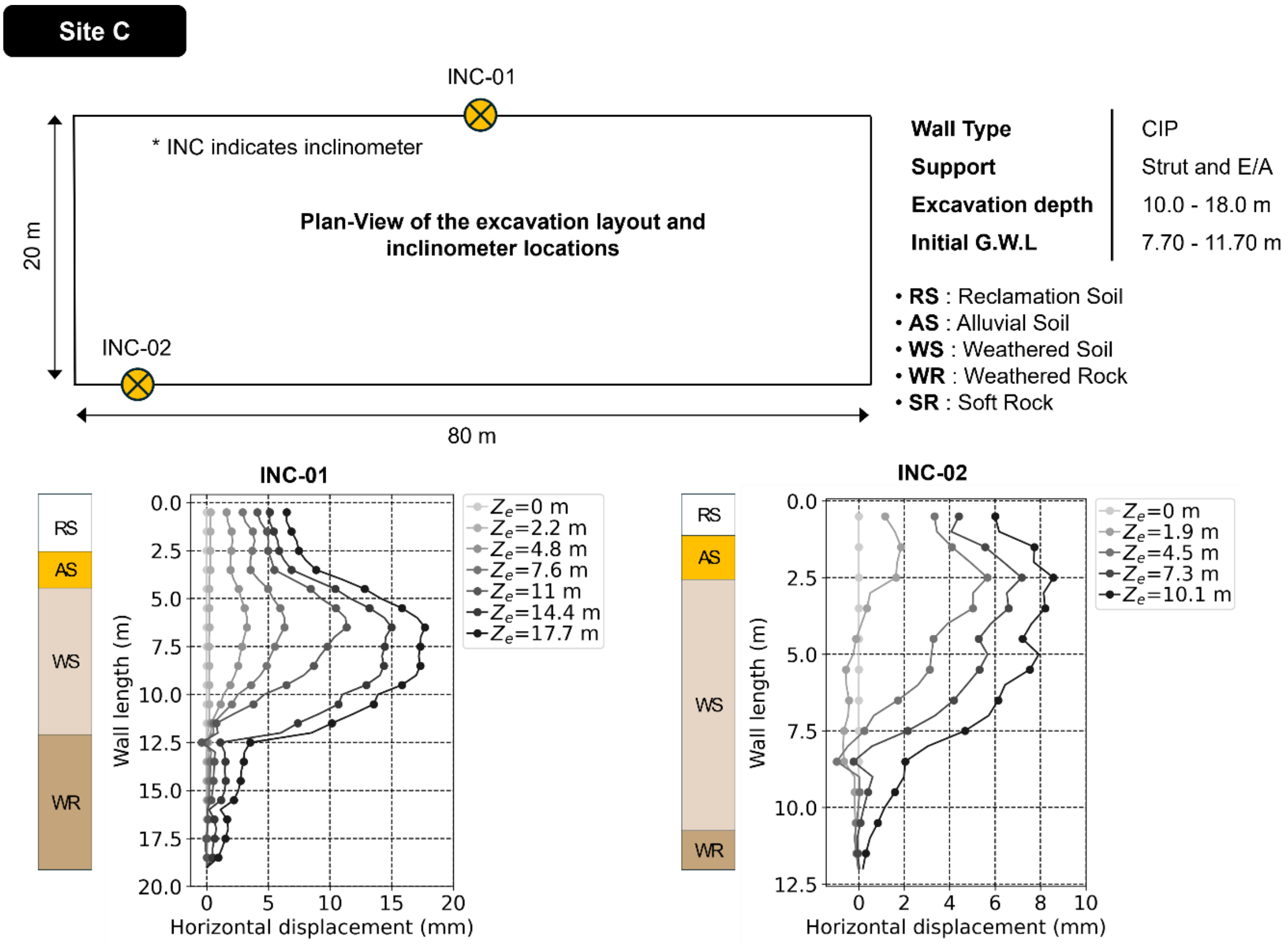

* Each curve represents the wall displacement profile at the excavation depth shown in the legend.

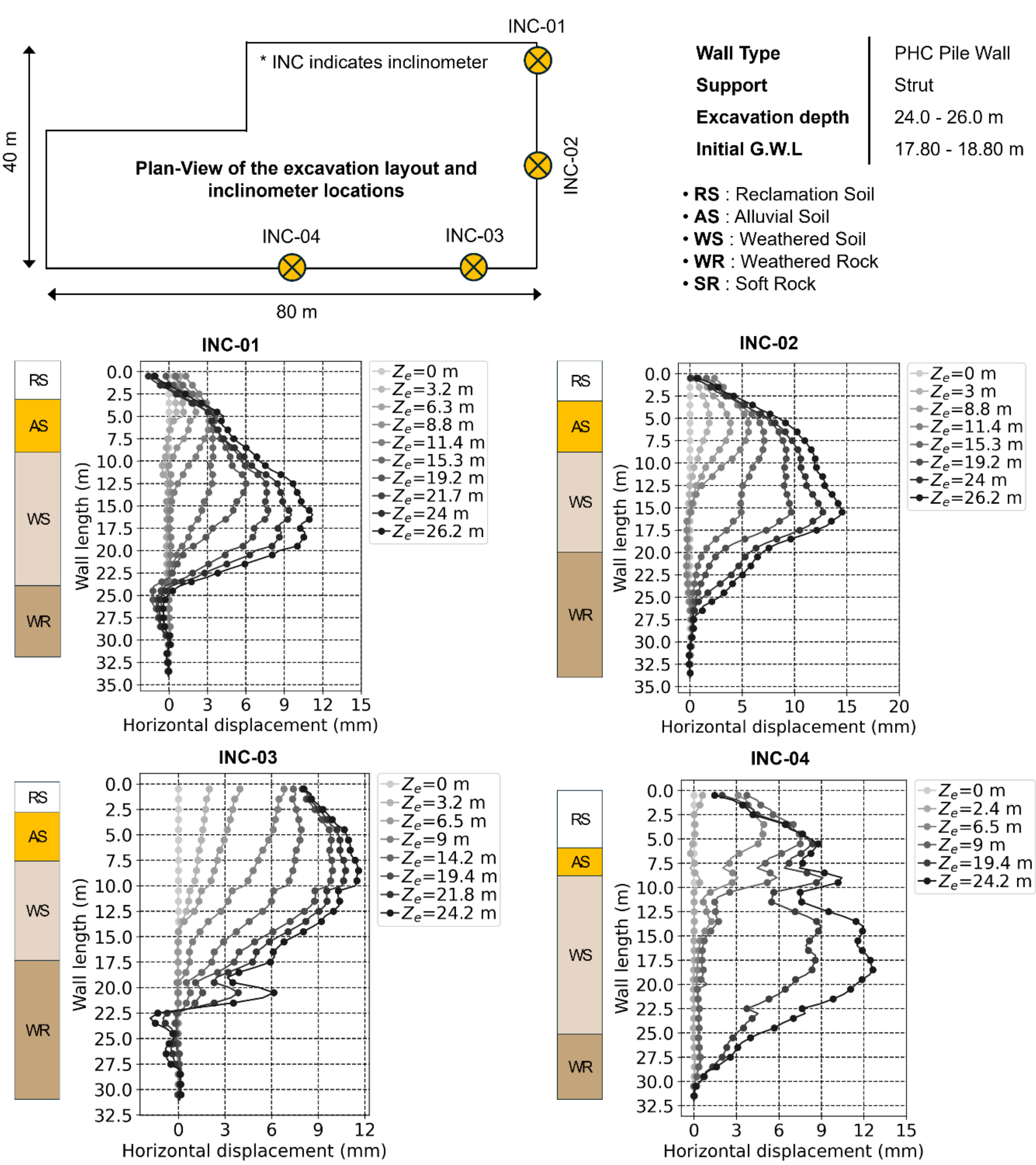

* Each curve represents the wall displacement profile at the excavation depth shown in the legend.

## Site E

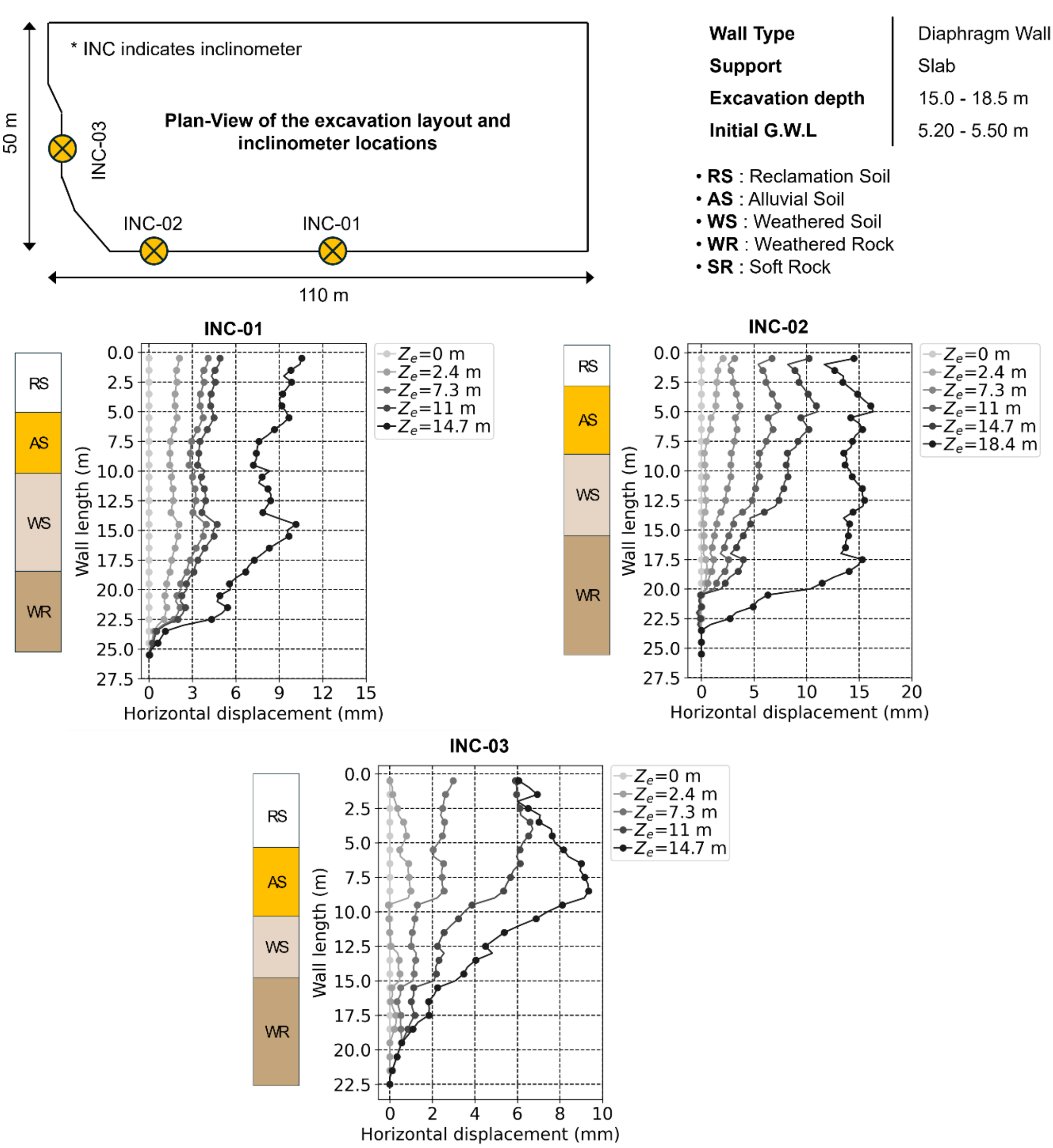

* Each curve represents the wall displacement profile at the excavation depth shown in the legend.

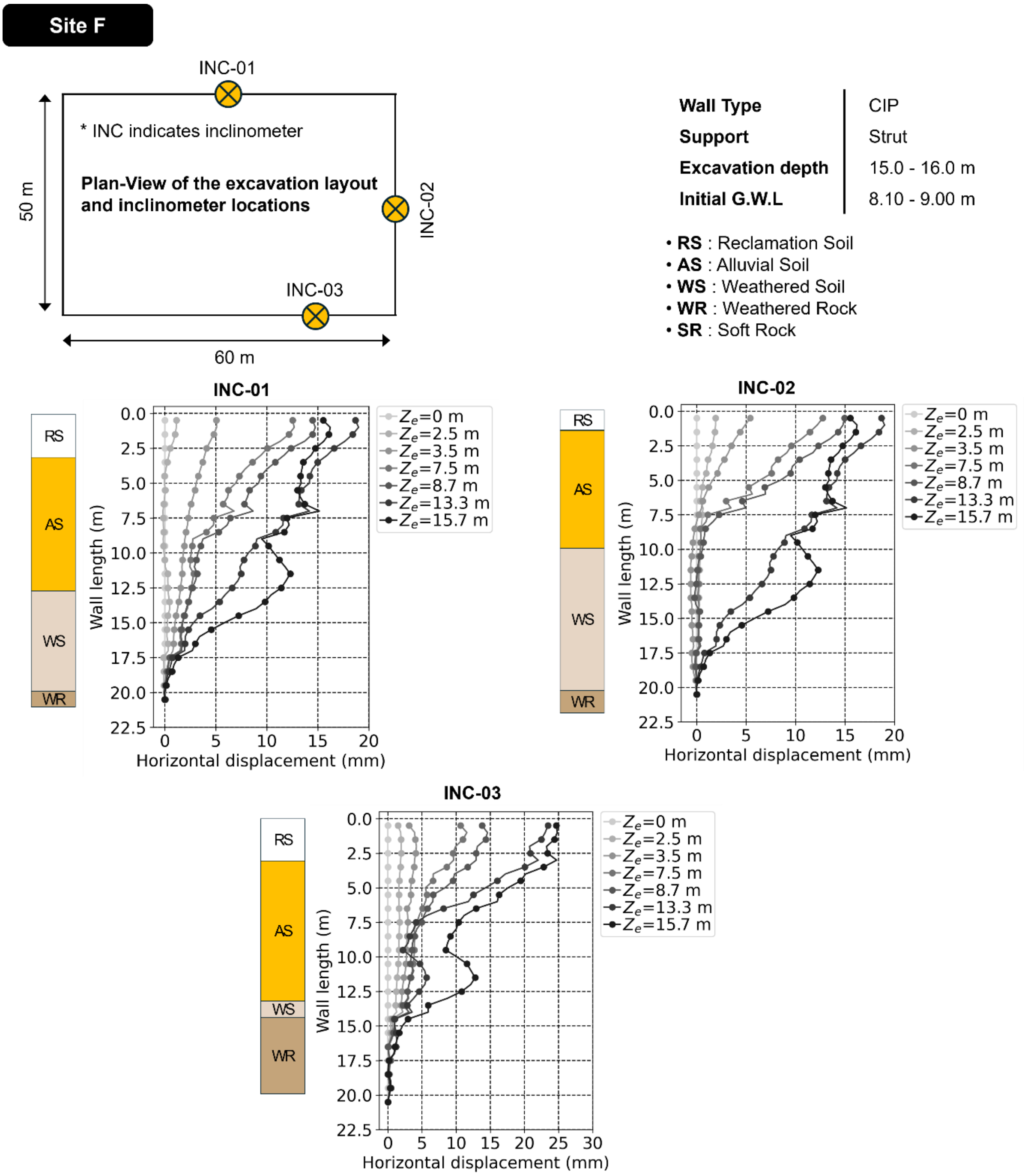

* Each curve represents the wall displacement profile at the excavation depth shown in the legend.

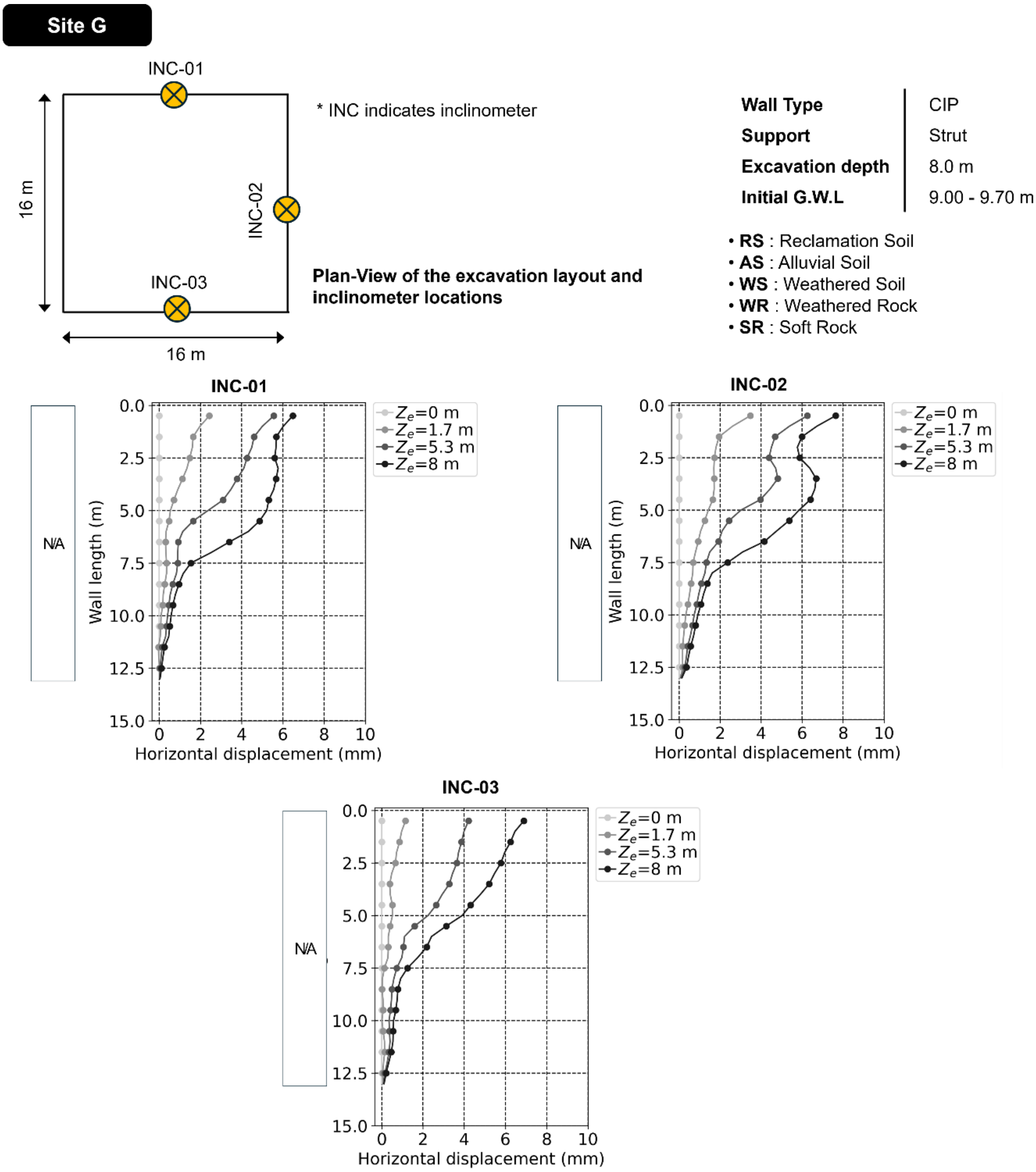

* Each curve represents the wall displacement profile at the excavation depth shown in the legend.

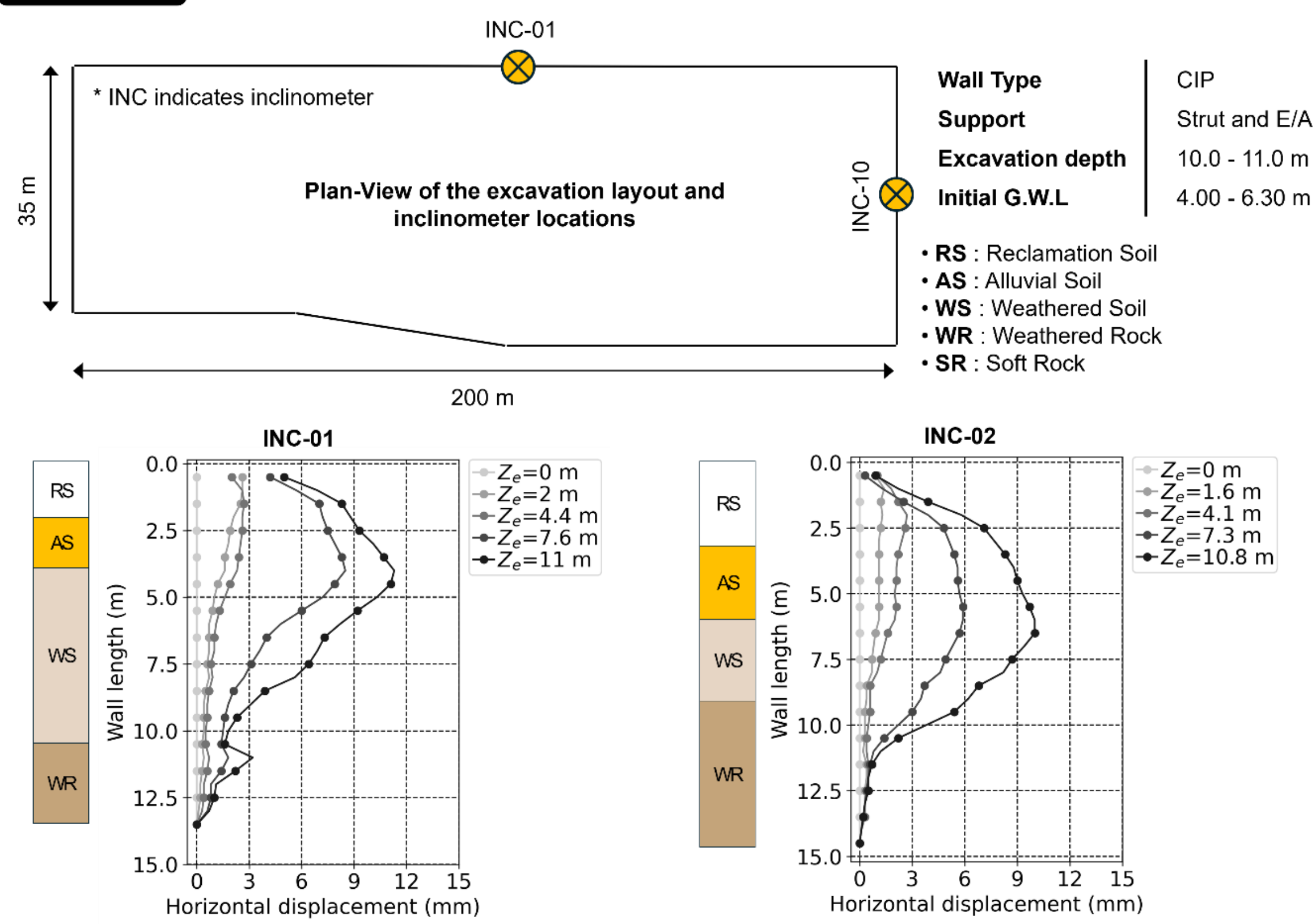

* Each curve represents the wall displacement profile at the excavation depth shown in the legend.

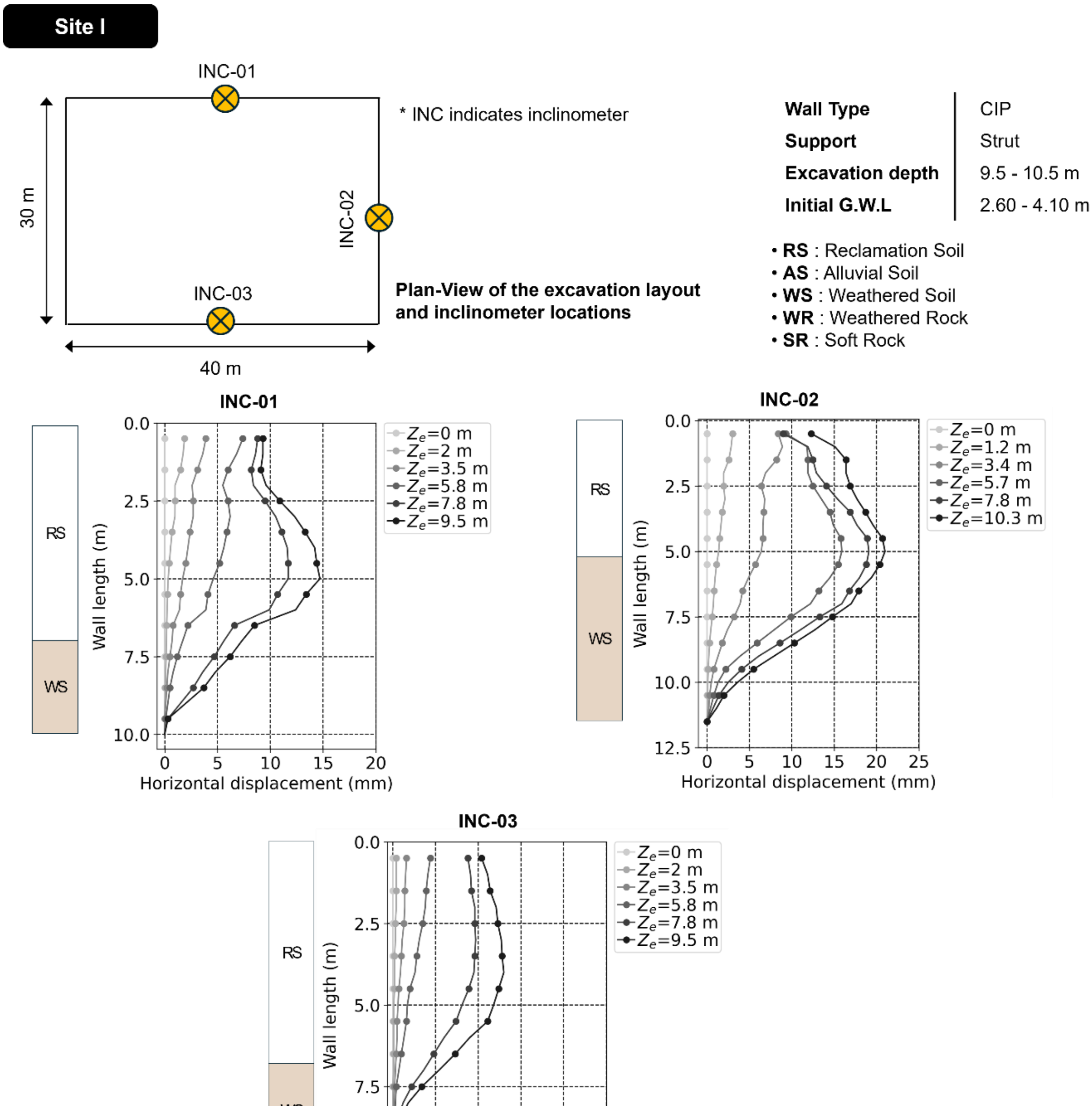

* Each curve represents the wall displacement profile at the excavation depth shown in the legend.

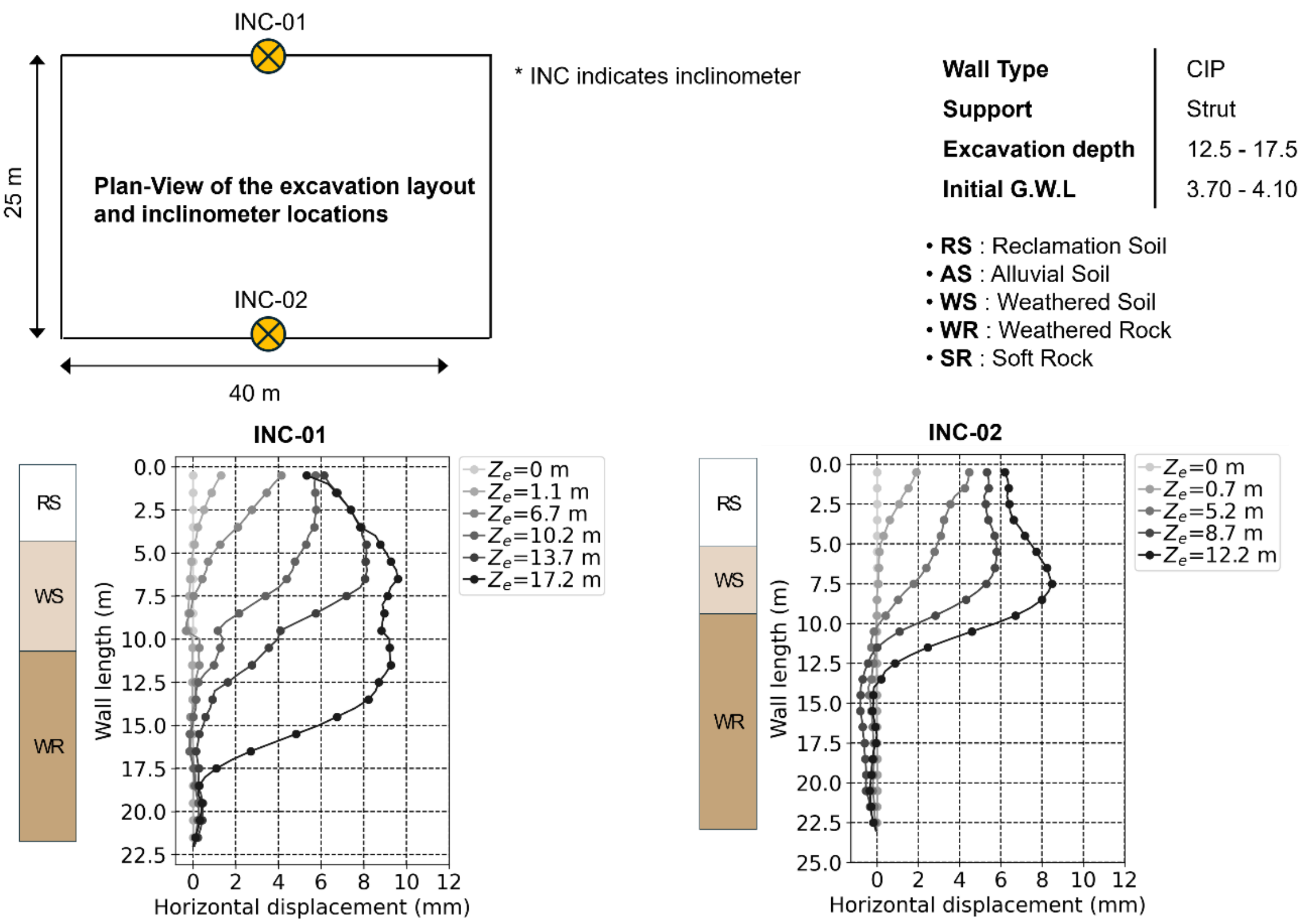

* Each curve represents the wall displacement profile at the excavation depth shown in the legend.

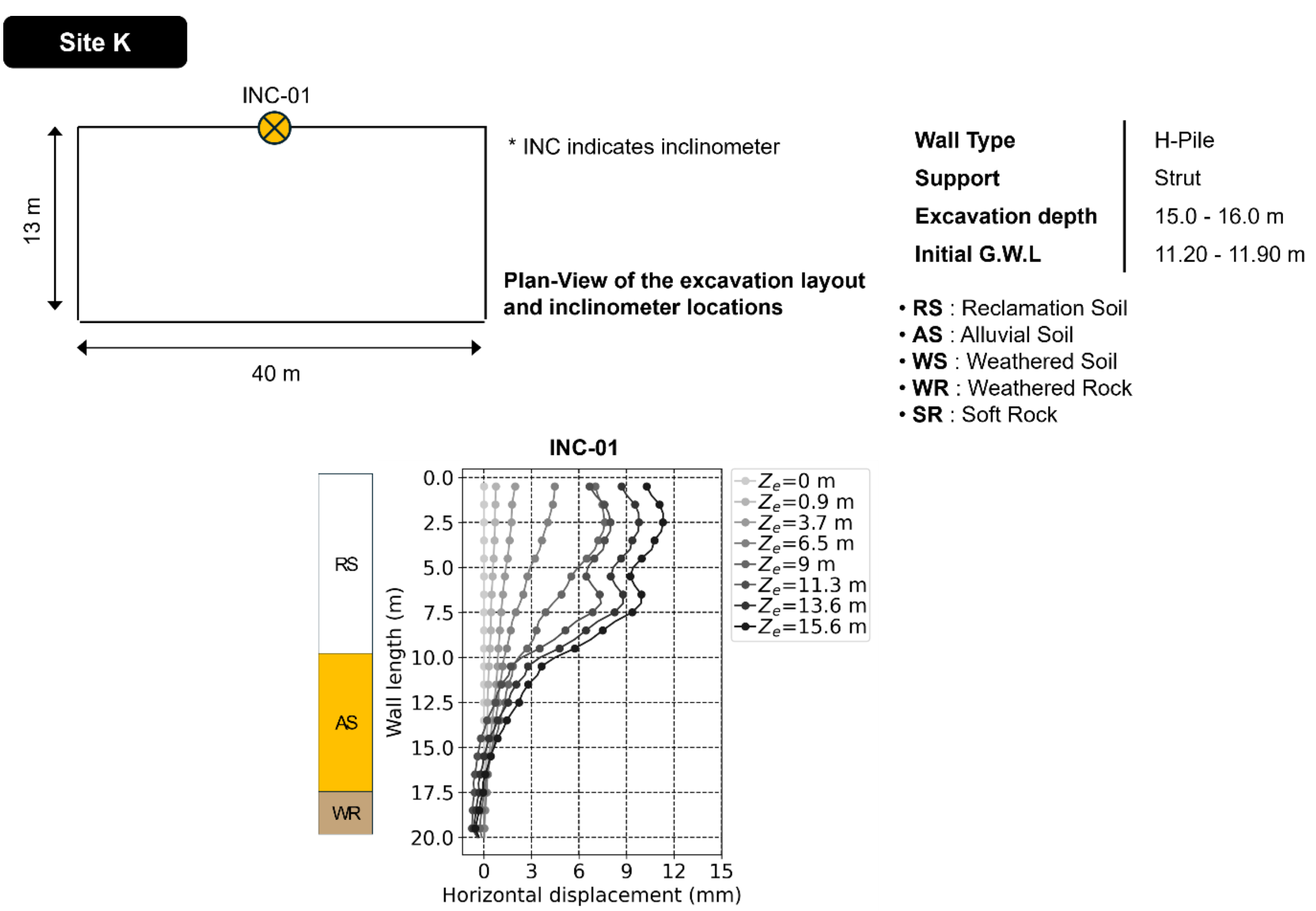